\documentclass[sigconf]{acmart}
\usepackage{enumitem}
\usepackage{threeparttable}
\usepackage{multirow}
\usepackage[font=normalsize]{subfig}
\AtBeginDocument{%
  \providecommand\BibTeX{{%
    \normalfont B\kern-0.5em{\scshape i\kern-0.25em b}\kern-0.8em\TeX}}}

\copyrightyear{2021}
\acmYear{2021}
\setcopyright{iw3c2w3}
\acmConference[WWW '21]{Proceedings of the Web Conference 2021}{April 19--23, 2021}{Ljubljana, Slovenia}
\acmBooktitle{Proceedings of the Web Conference 2021 (WWW '21), April 19--23, 2021, Ljubljana, Slovenia}
\acmPrice{}
\acmDOI{10.1145/3442381.3449950}
\acmISBN{978-1-4503-8312-7/21/04}

\begin{document}

\title{Mitigating Gender Bias in Captioning Systems}

\author{Ruixiang Tang, Mengnan Du, Yuening Li, Zirui Liu, Na Zou, Xia Hu}
\affiliation{
 \institution{Department of Computer Science and Engineering, Texas A\&M University} 
 \city{}
 }
\email{{rxtang, dumengnan, liyuening, tradigrada, nzou1, xiahu} @ tamu.edu}

\begin{abstract}
Image captioning has made substantial progress with huge supporting image collections sourced from the web. However, recent studies have pointed out that captioning datasets, such as COCO, contain gender bias found in web corpora. As a result, learning models could heavily rely on the learned priors and image context for gender identification, leading to incorrect or even offensive errors. To encourage models to learn correct gender features, we reorganize the COCO dataset and present two new splits COCO-GB V1 and V2 datasets where the train and test sets have different gender-context joint distribution. Models relying on contextual cues will suffer from  huge gender prediction errors on the anti-stereotypical test data. Benchmarking experiments reveal that most captioning models learn gender bias, leading to high gender prediction errors, especially for women. To alleviate the unwanted bias, we propose a new Guided Attention Image Captioning model (GAIC) which provides self-guidance on visual attention to encourage the model to capture correct gender visual evidence. Experimental results validate that GAIC can significantly reduce gender prediction errors with a competitive caption quality. Our codes and the designed benchmark datasets are available at \url{https://github.com/datamllab/Mitigating_Gender_Bias_In_Captioning_System}.
\end{abstract}



\maketitle

\section{Introduction}

Automatically understanding and describing visual contents is an important and challenging interdisciplinary research topic \cite{fang2015captions, hossain2019comprehensive, karpathy2015deep, mao2014deep, vinyals2015show, xu2015show}. Over the past years, thanks to the rapid development of deep learning and huge training data sourced from the web, captioning models have made substantial progress and even surpass humans with regard to several accuracy-based metrics \cite{vedantam2015cider}. Although a lot of efforts have been dedicated to improving the overall caption quality, few works consider the potential bias extensively existed among web sourced datasets. 

In this work, we investigate the gender bias problem and show that the widely used COCO dataset encodes gender stereotypes in web corpora. Firstly, the occurrence of men in image is significantly higher than women. More precisely, COCO \cite{lin2014microsoft} has an unbalanced 1:3 women to men ratio. Second, this gender disparity is even more obvious when considering the gender-context joint distribution. For example, most images about sport co-occur more frequently with men such as 90\% of surfboard images only contain male players. As a result, learning models may heavily rely on contextual cues to provide gender identification, e.g., always predicting a person as "woman" when the image is taken in the kitchen, which can lead to incorrect and even offensive gender predictions in domains where unbiased captions are required. The presence of strong priors in the dataset not only results in biased models but also makes it hard to detect the bias learned by models. Due to the i.i.d. nature of randomizing the split between train and test set \cite{geirhos2020shortcut, agrawal2018don}, biased models relying on incorrect visual evidence can still achieve competitive performance on the test set that has similar priors in the training set. This is problematic for validating progress in image captioning since it becomes unclear whether the improvements derive from learning correct visual features or empirical associations.

We present two new COCO splits: COCO-GB V1 and V2 datasets, so as to reveal gender bias learned by captioning models. These two splits are created by reorganizing data distribution such that for each gender, the distribution of image context is very different between train and test set. Specifically, COCO-GB V1 is designed to measure gender bias in existing models systematically. COCO-GB V2 is created to further assess the capabilities of models in  gender bias mitigation. Our hypothesis is that models relying on image context to provide gender identification will suffer from a huge gender prediction error on the anti-stereotypical test dataset. In this way, we can reveal the mismatch between human-intended and model-captured features and quantify gender bias based on the gender prediction outcomes on our new settings. 

 Equipped with our new benchmark datasets, we evaluate several widely used captioning models.
 The key observation is that most models learn or even exaggerate gender bias in the training data, which causes high gender prediction errors, especially for women. Another important finding is that commonly used evaluation metrics, such as BLEU \cite{papineni2002bleu} and CIDEr \cite{vedantam2015cider}, mainly focus on the overall caption quality and are not sensitive to gender error. Our experimental results show that a baseline model achieves competitive caption quality scores on these metrics even when it has misclassified 27\% of images with women into men. The benchmarking experiments indicate that without extra regularization, captioning models are prone to replicate bias in the dataset. The “high” performance achieved by existing models should be revisited. 


\begin{figure*}[t]
    \centering
    \includegraphics[width=0.9\linewidth]{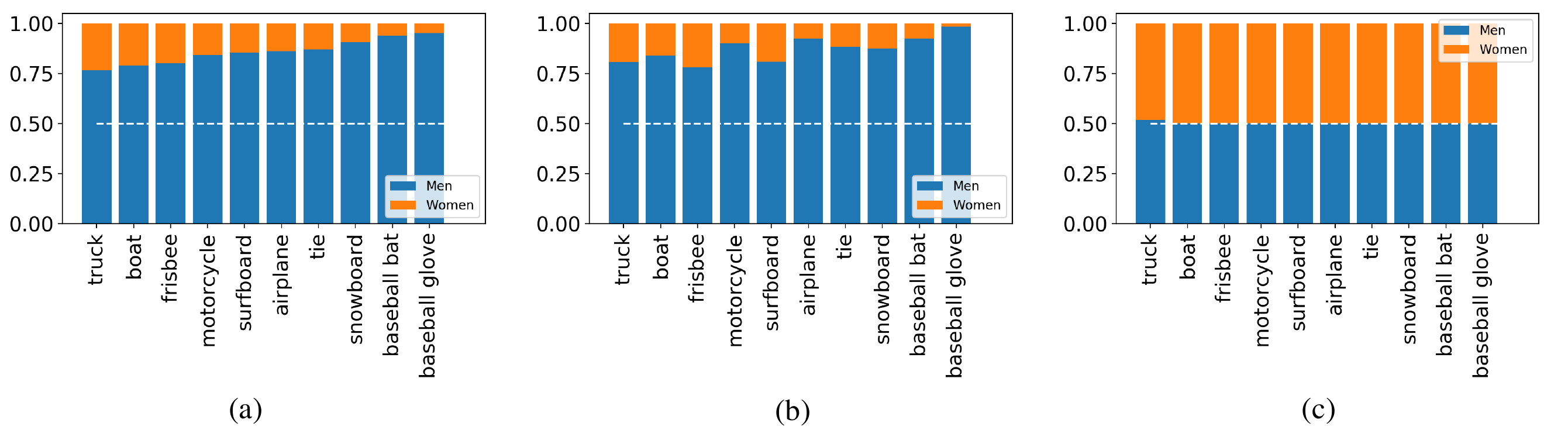}
    \vspace{-10pt}
    \caption{ We select several objects in COCO dataset and show their gender distribution in (a) training data, (b) testing data and (c) COCO-GB V1 secret test data. Our key observation is that there is a significant bias in training set, e.g., more than 90\% objects have higher probability to co-occur with men. We can find that similar bias also exists in the original test set, while the secret test dataset from COCO-GB V1 has a balanced gender-context distribution.} 
    \label{fig:data_distribution}
\end{figure*}

In addition to reveal gender bias in learning models, we also seek to mitigate gender bias by guiding the model to capture correct gender features. From the data perspective, a straightforward solution is to train the model on a dataset with equal training samples for each gender. Unfortunately, experimental results indicate that simply balancing image numbers has limited improvement in bias mitigation. From the training perspective, an alternative approach is to increase the loss weight of gender words, which also doesn't achieve a satisfactory result. 

To overcome the unwanted bias, we propose a new Guided Attention Image Captioning model (GAIC). GAIC has two complementary streams to encourage the model to explore correct gender features by self-guided supervision. The new training paradigm encourages the model to provide correct gender identification with high confidence when gender evidence in image is obvious. When gender evidence is vague or occluded, GAIC tends to describe the person with gender neutral words, such as "person" and "people." In addition to self-supervised learning, we also consider the semi-supervised scenarios where a small amount of extra supervision is accessible (e.g., person segmentation masks). GAIC training pipeline can seamlessly add extra supervision to accelerate the self-exploration process and further improve gender prediction accuracy. The proposed training paradigm of GAIC is model-agnostic and can be easily applied to various captioning models. Experimental results validate that GAIC can significantly reduce gender prediction errors on our new benchmark datasets, with a competitive caption quality. Results of visualizing attention further prove that GAIC is more inclined to adopt the person's appearance for gender identification. We conclude our contributions as follows:
\begin{itemize}[leftmargin=*]
\setlength\itemsep{0em}
\item Through reorganizing data distribution, we present two new splits COCO-GB V1 and COCO-GB V2 datasets to quantify gender bias learned by captioning models. 
\item We benchmark several captioning baselines on COCO-GB V1. Experimental results reveal that most models have gender bias problems, leading to high gender prediction errors. 
\item We propose a new Guided Attention Image Captioning model (GAIC). By self-guided exploration, the model learns to focus on correct gender evidence for gender identification. 
\item Experimental results demonstrate that the proposed GAIC model can significantly reduce gender prediction errors while at the same time preserving a competitive caption quality.
\end{itemize}

The rest of this paper is organized as follows. \textbf{Section~2} discusses the procedures of dataset creation. \textbf{Section~3} presents the benchmarking experiment results on COCO-GB V1. \textbf{Section~4} introduces the proposed GAIC and GAIC$_{es}$ model. \textbf{Section~5} includes experimental results to verify effectiveness of the proposed method.


\newcommand{\tabincell}[2]{\begin{tabular}{@{}#1@{}}#2\end{tabular}}
\begin{table*}[t]
\begin{threeparttable}
\centering
\begin{tabular}{ccccccccccc}
\toprule
\multicolumn{1}{c|}{\multirow{2}{*}{Model}} & \multicolumn{1}{c|}{\multirow{2}{*}{BLEU-4}} & \multicolumn{1}{c|}{\multirow{2}{*}{CIEDr}} & \multicolumn{1}{c|}{\multirow{2}{*}{METEOR}} &
\multicolumn{1}{c|}{\multirow{2}{*}{\tabincell{c}{Gender Error \\ (Original/Secret)}}} &
\multicolumn{3}{c|}{Women} & \multicolumn{3}{c}{Men} \\
\cline{6-11} 
\multicolumn{1}{c|}{} & \multicolumn{1}{l|}{} & \multicolumn{1}{l|}{} & \multicolumn{1}{l|}{} &
\multicolumn{1}{l|}{} & \multicolumn{1}{l|}{correct} & \multicolumn{1}{l|}{wrong} & \multicolumn{1}{l|}{neutral} & \multicolumn{1}{l|}{correct} & \multicolumn{1}{l|}{wrong} & neutral \\ \hline
FC  & 31.4 & 95.8 & 24.9 & 9.7 / 12.6 & 60.5 & 14.8 & 24.7 & 64.3 & 10.3 & 25.4    \\
LRCN  & 30.0 & 90.8 & 23.9 & 11.3 / 14.0 & 61.7 & 16.8 & 21.6 & 64.7 & 11.2 & 24.0   \\
Att & 31.0 & 95.1 & 24.8 & 6.7 / 15.6 & 53.2 & 25.1 & 22.7 & 62.7 & 4.4 & 32.9    \\
AdapAtt & 31.1 & 98.2 & 25.2 & 5.5 / 15.3 & 47.9 & 26.8 & 27.4 & \textbf{75.7} & \textbf{3.8} & 20.5    \\
Att2in & 32.8 & 102.0 & 23.3 & 8.3 / 11.4 & 61.5 & 16.3 & 22.1 & 70.2 & 6.5 & 23.3    \\
\hline
TopDown${\dagger}$ & 34.6 & 107.6 & 26.7 & 7.8 / 8.2 & 65.5 & \textbf{9.0} & 25.5 & 63.5 & 7.4 & 29.0  \\
NBT${\dagger}$ & 34.1 & 105.1 & 26.2 & 5.1 / 6.8 & \textbf{72.3} & 9.3 & 18.3 & 77.6 & 4.3 & 18.0    \\
\hline
FC$^{*}$ & 33.4 & 103.9 & 25.0 & 9.1 / 13.8 & 61.6 & 20.7 & 17.7 & 71.9 & 6.8 & 21.3    \\
LRCN$^{*}$  & 29.4 & 93.0 & 23.5 & 12.1 / 13.4 & 68.1 & 11.3 & 20.6 & 60.6 & 15.5 & 23.9   \\
Att2in$^{*}$  & 33.6 & 106.7 & 25.7 & 9.3 / 12.4 & 61.8 & 19.7 & 18.5 & 73.8 & 6.2 & 20.0    \\
TopDown${\dagger}*$  & \textbf{34.9} & \textbf{117.2} & \textbf{27.0} & 9.1 / 11.3 & 69.0 & 15.1 & 15.8 & 73.3 & 7.5 & 19.1  \\
\bottomrule
\end{tabular}
\caption{Gender bias analysis on COCO-GB V1 split. We utilize BLEU-4(B-4), CIEDr(C) and METEOR(M) to evaluate captions qualities, all results are generated with beam size 5. Caption quality is obtained from test dataset, and gender bias is evaluated on COCO-GB V1 secret test dataset. ${\dagger}$ denotes the models that utilize extra grounded information from the Faster R-CNN network. $(^{*})$ denotes the models that are trained with self-critical loss.}
\label{tab:baseline_gender_bias}
\end{threeparttable}
\vspace{-20pt}
\end{table*}

\section{COCO-GB V1 and V2}
 In this section, we present two new splits COCO-GB V1 and V2 to reveal gender bias in learning models. 
\label{sec:3}

    \vspace{2pt}
    \noindent\textbf{Gender Labeling:} Gender identification is not an explicit task for image captioning, hence COCO dataset doesn't particularly label the person's gender. Our first step is to annotate the gender of people in the images. Because many images do not have a clear human face, existing face recognition systems cannot be directly used to annotate the gender. Alternatively, we take inspiration from \cite{hendricks2018women} and make use of the five human-annotated captions available for each image to label the person's gender. Images are labeled as "women" if at least one of the five descriptions contain the female gender words and do not include male gender words. Similarly, images are labeled as "men" if at least one of the five descriptions contain the male gender words and do not include female gender words. Images mentioned both "man" and "woman" are discarded. To improve the label accuracy, we only consider images containing one person and remove images with multiple people.
    
    We then conduct human evaluations on our gender annotations with majority voting from 3 human evaluators. Each evaluator validates a total of 400 test samples from the labeled data, with 200 randomly sampled from each gender. Image is manually labeled from "women," "men," "women \& men" (if women and men are included in a single picture) and "discard" (no human appears in the image or gender is indistinguishable). Results show that our gender annotation achieves a high precision, 92\% for women, and 95\% for men. By analyzing the falsely annotated cases, we find that the gender evidence for these images usually is too vague to distinguish, which makes gender identification far more challenging. (list of gender words, gender annotation examples are shown in Sec.~\ref{sec:label gender}). 
    
    \vspace{2pt}
    \noindent\textbf{COCO-GB V1:} COCO-GB V1 is created for systematically evaluating gender bias in existing models. For ease of evaluation, we construct the dataset based on a widely used split proposed by Karpathy et al \cite{karpathy2015deep} (Karpathy split), and collect a gender-balanced secret test dataset from its test split. In this way, captioning models trained on Karpathy split can directly evaluate their gender bias on this secret test set without retraining. Previous work \cite{hendricks2018women} also introduces a small dataset with 1:1 women to men ratio to evaluate gender error. However, it does not consider the bias caused by gender-context co-occurrence. In comparison, our proposed secret test dataset reduces the bias in the gender-context distribution and thus can reflect the real performance. To achieve it, we calculate the gender bias rate towards men for 80 COCO object categories by metrics proposed by work \cite{zhao2017men}:
    \begin{equation}
        \frac{ count(object,men)}{ count(object,men) + count(object, women)},
        \label{eq:gender bias}
    \end{equation}
    where \textit{men} and \textit{women} refer to images labeled as "men" and "women", $count(object, gender)$ refers to the co-occurrence counts of the object (e.g., Motorcycle and Frisbee) with certain gender. For an object, a bias ratio of 0.5 represents that women and men have an equal probability of co-occurring with the object. High/low bias ratio represents that the object is more likely to co-occur with men/women. We obtain the bias ratio of 80 COCO objects in train and test data using Eq.~\ref{eq:gender bias} (examples are shown in Fig.~\ref{fig:data_distribution}). The results show that objects in training data have an average bias ratio of 0.65, and 90\% of objects are more likely to co-occur with men. A similar bias is also found in the test dataset. We note that the demonstration of bias in COCO is a refection of social bias captured in web. 
    
   Evaluating models on the biased test dataset is problematic since models can make use of the contextual cues existed in training data to provide "correct" gender identification. To construct an unbiased test dataset, we utilize a greedy algorithm to select a secret test dataset from the original test split so that each object category has a nearly equal probability  of appearing with women or men. The created secret test dataset has 500 images for each gender, and the average bias ratio drops from 0.65 to 0.52. We then validate several existing captioning models on the COCO-GB V1 secret dataset.
    
    \vspace{3pt}
    \noindent\textbf{COCO-GB V2:} This dataset is designed to further assess the model robustness when exposed to novel gender-context pairs at test time. To create the new split, we first sort 80 object categories in COCO dataset according to their gender bias ratios. Unlike creating a balanced test dataset in COCO-GB V1, we start from the most biased object and greedily add selected data into the test set. As a result, the difference has been dramatically enlarged between the gender-context joint distribution of train and test data. Our sampling algorithm guarantees that there are sufficient images from each category during training, but at test time model will face novel compositions of gender-context pairs, e.g., women with the skateboard. The final split has 118,062/5,000/10,000 images in train/val/test respectively (complete gender-context distribution of COCO, COCO-GB V1 and COCO-GB V2 are shown in Sec.~\ref{sec:coco_gender}). 

\section{Benchmarking Existing Models}

In this section, we benchmark several baselines on the COCO-GB V1 dataset. Models are trained on Karpathy split. We evaluate the caption quality on the original test data and report gender prediction results on both original and COCO-GB V1 test dataset.

\vspace{2pt}
\noindent\textbf{Baselines:} We compare gender bias across a wide range of models. From the model architecture perspective, we consider models both with and without attention module where "attention" refers to models learn to pay attention to different image regions for caption word generation \cite{rohrbach2018object}. For non-attention models, we consider 
\textbf{FC} \cite{rennie2017self} which initializes LSTM with features extracted directly by CNN, and \textbf{LRCN} \cite{donahue2015long} which leverages visual information at each time
step. For attention models, we select \textbf{Att} \cite{xu2015show}, which firstly applies visual attention mechanism in caption generation. \textbf{AdapAtt} \cite{lu2017knowing}, which automatically determines when and where to put visual attention. \textbf{Att2in} \cite{rennie2017self}, which modifies the architecture of Att, and inputs the attention features only to the cell node of the LSTM. Besides, we also consider models that utilize extra visual grounding information. \textbf{TopDown} \cite{anderson2018bottom} proposes a novel “top-down attention” mechanism based on Faster R-CNN model \cite{ren2015faster}. \textbf{NBT} \cite{lu2018neural} generates the sentence “template” with slot locations, and fill slots by an object detection model. All captioning models are end-to-end trainable and use LSTM \cite{hochreiter1997long} to output caption text.

\vspace{2pt}
\noindent\textbf{Learning Objective and Implementation} : We use the cross-entropy loss as the objective function. Besides, FC, LRCN, Att2in, and TopDown are also trained using self-critical loss \cite{rennie2017self} that uses reinforcement learning to optimize the non-differentiable CIDEr metric. For a fair comparison, all baselines utilize visual features extracted from the ResNet-101 network, NBT and TopDown model
adopt extra grounded information from the Faster R-CNN model.

\vspace{2pt}
\noindent\textbf{Evaluation Metrics and Results Analysis:} In Tab.~1, we report the caption quality as well as gender prediction performance. To evaluate caption quality, we adopt several commonly used metrics, such as BLEU, CIEDr and METOR \cite{denkowski2011meteor}, measure the similarity between machine-generated captions and human-provided annotations. For gender prediction, results are grouped into three circumstances: correct, wrong, and neutral (no gender-specific words are generated). Because of the sensitive nature of prediction for protected attributes (gender words in this work), we emphasize the importance of low error rate \cite{hendricks2018women}. Also, we encourage the model to use gender neutral words in cases where the model has low confidence. Based on the results, we reach the following conclusions.
\begin{itemize}[leftmargin=*]
\setlength\itemsep{0em}
\item  In Tab.~\ref{tab:baseline_gender_bias}, Gender Error measures the error rates when describing women and men, we find a notable performance gap between the original test dataset (average error rate of 7.7\%) and COCO-GB v1 secret test dataset (average error rate of 12.1\%), which proves that the original test split indeed underestimates the gender bias learned by models. Tab.~\ref{tab:baseline_gender_bias} also reports the outcome distribution for each gender on the secret test. We observe that the error rate of women is substantially higher than men, the average error rate of all models for women and men is 16.7\% and 7.7\%, respectively. An interesting finding is that models with the attention mechanism have much higher errors for women (average of 22.6\%) compared to non-attention models (average of 15.8\%). One possible explanation is that the attention mechanism strengthens models' ability to capture visual concepts; on the other hand, also makes models prone to learn contextual bias. Another finding is that models using extra visual grounding information, such as NBT and TopDown, have a much lower error rate, especially for women (average of 9.1\%), which indicates that the extra visual features provided by Faster R-CNN model are unbiased and useful for gender prediction. On the flip side, these models' gender prediction accuracy will highly depend on the features provided by the object detection models.

\vspace{2pt}
\item Models with a high gender error rate can still get competitive caption quality scores. For example, AdapAtt model obtains a decent caption quality performance on three evaluation metrics but at the same time, has the highest women error rate across all models, which misclassifies 26.8\% of images labeled as women into men. The experimental results demonstrate that existing evaluation metrics mainly focus on the overall caption quality and are not sensitive to the gender word error. New evaluation metrics that can effectively lead to high caption quality and low gender prediction errors are still lacking.

\vspace{2pt}
\item Self-critical loss improves the models' overall caption quality but also significantly amplifies the gender error rate. CIEDr metric obtains an average improvement of 6.2\% after training with self-critical loss. However, the error rate of women in FC, Att2in, and TopDown model increases by an average of 5.1\% and  the error rate of men in LRCN model increases by 4.2\%. This result suggests once again that the improvement of overall caption quality cannot guarantee a high accuracy rate of gender identification.
\end{itemize}
\label{sec:benchmarking}
\begin{figure*}[t]
    \centering
    \includegraphics[width=0.95\linewidth]{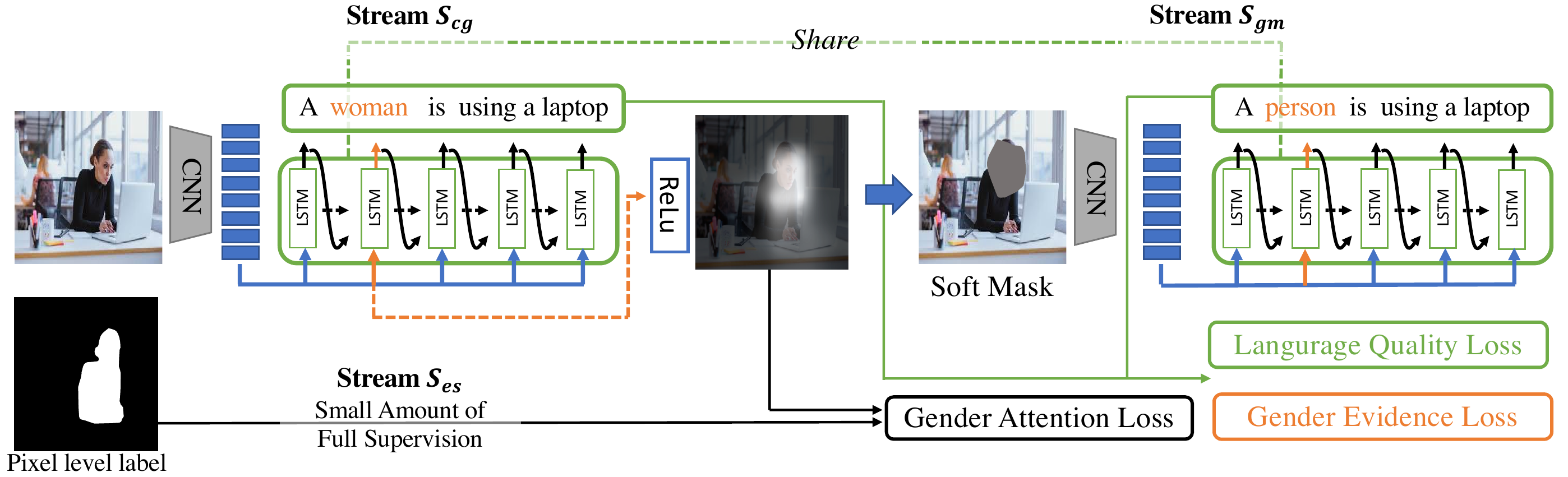}
    \vspace{-10pt}
    \caption{GAIC has two streams of networks that share parameters. Stream S$_{cg}$ finds out regions that help model to classify the gender, and Stream S$_{gm}$ tries to make sure all selected regions are correct gender evidence features. The attention map is online generated and two streams are trained by the Language Quality Loss and Gender Evidence Loss jointly. GAIC$_{es}$ model seamlessly adds a small amount of extra supervision to further refine model attention which denotes as S$_{es}$.}
    \label{fig:model_pipeline}
    \vspace{-5pt}
\end{figure*}

\section{The Proposed GAIC Framework}
Attention mechanism has been widely used in image captioning task, which significantly improves the caption quality \cite{anderson2018bottom, lu2017knowing, xu2015show}. However, our benchmarking experiment shows that when no explicit enforcement is made to infer the gender by only "looking" at the person, models with the attention mechanism are more prone to predict the gender using the context in the image. To overcome the unwanted bias, we propose the Guided Attention Image Captioning model (GAIC) which provides self-guidance on visual attention to encourage the model to utilize correct gender features. More specifically, GAIC considers a situation with no grounded supervision in which the model explores the correct gender evidence by model itself. In this way, the training paradigm of GAIC is model-agnostic and can be easily applied to various captioning models. In addition to self-exploration, we also consider the semi-supervised scenario that uses a small amount of extra supervision to further control attention learning process.

\subsection{Caption Generation with Visual Attention}

We start by briefly describing the captioning model with attention mechanism. Given an image $I$ and the corresponding caption $y=\{y_1,...,y_T\}$, the objective of an encoder-decoder image captioning model is to maximize the following loss function:
\begin{equation}
    {\arg\max\limits_\theta} \sum\limits_{(I,y)} \log p(y|I;\theta) = \sum\limits_{t=1}^{T}\log p(y_t|y_1,...,y_{(t-1)},I),
\end{equation}
where $\theta$ is the trainable parameters of the captioning model. We utilize chain rule to decompose the joint probability distribution into ordered conditionals. Then a recurrent neural network such as LSTM predicts each conditional probability as follows:
\begin{equation}
\log p(y_t|y_1,...,y_{t-1}, I) = f(h_t, c_t),
\end{equation}
where $f$ is a nonlinear function that predicts the probability of $y_t$. $h_t$ is the hidden state of LSTM at $t$ steps. $c_t$ is the visual context vector extracted from image $I$ for predicting $t^{th}$ caption word $y_t$. We follow the work \cite{xu2015show} to compute $c_t$ by:
\begin{equation}
c_t = \sum\limits_{k=1}^{K}\alpha_{t,k}v_k,
\end{equation}
where $V=\{v_1,...,v_k\}$,$v_i \in \mathbb{R}^d$ is a set of image features extracted from last convolutional layer of a CNN network. $\alpha=\{\alpha_1,...,\alpha_T\}$, $ \alpha_t \in \mathbb{R}^K$ is the attention weight over features in $V$. Based on the attention distribution $\alpha_t$, we extract useful information from image features to obtain context vector $c_t$. By visualizing the attention weight $\alpha_t$ learned by the model, we are able to analyze on which region of the image does the network focus  when generating the $t^{th}$ caption word. On the other hand, we can also modify the generated captions by adding regularization on the $\alpha$.
\label{sec:atteniton caption model}

\subsection{Self-Guidance on Gender Word Attention}
To achieve the goal of self-supervision on network attention, we design a two-stream training pipeline. As shown in Fig.~\ref{fig:model_pipeline}, GAIC includes caption generation stream $S_{cg}$ and gender evidence mining stream $S_{gm}$. The two streams share parameters with each other. The purpose of stream $S_{cg}$ is to output high-quality captions and generate attention maps of gender words. The second stream $S_{gm}$ will force the attention generated by $S_{cg}$ to be focused on the correct regions in the image. The two complementary streams guide the attention regions of gender words to be focused on correct features, such as human appearance, and keep away from contextual cues like skateboard and laptop. 

For stream $S_{cg}$, given an image $I$ and ground truth caption $y=\{y_1,...,y_T\}$ , the model outputs a caption supervised by the Language Quality Loss $\mathcal{L}_{lq}$, e.g., the commonly used cross-entropy loss. In addition, we obtain the attention values $\alpha = \{\alpha_{t} \mid t\in[1,T]\}$ for each caption word where $\alpha$ can be directly obtained from models with attention mechanism like we mentioned in Sec.~\ref{sec:atteniton caption model}. For non-attention models, $\alpha$ can be approximated by the post-hoc interpretation methods, such as Grad-CAM \cite{selvaraju2017grad} and Saliency Map \cite{simonyan2014deep}.  In this work, we obtain $\alpha$ directly from an attention model and only consider the attention map of gender words which is denoted as $\alpha^g$. Such that $\alpha^g$ 
enables us to visualize the evidence for generating gender words. We then utilze $\alpha^g$ to generate a soft mask that is applied on the original input image. In this way, we obtain $I^{*g}$ using Eq.~\ref{eq:mask_ig} where $I^{*g}$ represents the image regions apart from the network's attention for gender identification.
\begin{equation}
I^{*g} = I-(I \odot T(\alpha^g)),
\label{eq:mask_ig}
\end{equation}
where $\odot$ denotes the element-wise multiplication. $T$ is a masking function built on a thresholding operation. To make the masking process derivable, we follow the work \cite{li2018tell} and use Sigmoid function as an approximation defined in Eq.~\ref{eq:sigmoid_attention}.
\begin{equation}
T(\alpha^g) = \frac{1}{1+\exp(-w(\alpha^g-E))},
\label{eq:sigmoid_attention}
\end{equation}
where $E$ is a matrix that all element equal to a threshold value $\sigma$. Scale parameter $w$ ensures $T(\alpha^g_i)$ approximately equals to 1 when  $T(\alpha^g_i)$ is larger than $\sigma$ and to 0 otherwise. 

We then feed $I^{*g}$ into the second stream $S_{gm}$ and output a new caption. Since $I^{*g}$ already removes gender related features learned by the model, we correspondingly replace the gender words in ground truth caption $y$ into gender neutral words according to the following replacement rules: 
\begin{multline}
    woman/man (etc) \rightarrow person, ~women/men (etc)
    \rightarrow people,\\
    boy/girl (etc) \rightarrow child, ~boys/girls (etc) \rightarrow children.    
\end{multline}
We retrain the model supervised by the new ground truth caption. The loss in $S_{gm}$ is denoted as Gender Evidence Loss $\mathcal{L}_{ge}$, which is defined as follows:
\begin{multline}
    \mathcal{L}_{ge} = {\arg\max\limits_\theta} \sum\limits_{t=1}^{T}\log p(y_t|y_1,...,y_{t-1}),I^{*g}),\\ y_t \Leftarrow g_n~if~y_t \in g_w \cup g_m,
\end{multline}
where $y_t$ denotes the $t^{th}$ caption word, $g_n$, $g_w$ and $g_m$ represent neutral, female and male gender words, respectively. Loss $\mathcal{L}_{ge}$ can be minimized only when models focus on correct gender features such as human appearance. For a biased model relying on the contextual cues for gender identification, $I^{*g}$ generated by $S_{cg}$ will remove the biased context, e.g., a laptop. As a result, the stream $S_{gm}$ will generate a low-quality caption because of missing the important context features and produce a high loss value for $\mathcal{L}_{ge}$. In such a way, stream $S_{gm}$ forces the stream $S_{cg}$ to capture the correct gender evidence. Besides learning proper gender features, $\mathcal{L}_{ge}$ also encourages the model to predict gender cautiously and use gender neutral words when gender evidence is vague in the image like $I^{*g}$ (we show an example in Fig.~\ref{fig:model_pipeline}). 
Finally, we combine $\mathcal{L}_{lq}$ and $\mathcal{L}_{ge}$ as the Self-Guidance Gender Discrimination Loss:
\begin{equation}
    \mathcal{L}_{self} = \mathcal{L}_{lq} + \mu\mathcal{L}_{ge},
\end{equation}
where $\mu$ is a weighting parameter to balance two streams. With the joint optimization of $\mathcal{L}_{self}$, the model learns to generate high quality captions as well as focus on correct visual features that contribute to gender recognition.

\subsection{Integrating with Extra Supervision} 
In addition to self-exploration training, we also consider the semi-supervised scenarios where a small amount of extra supervision is added to accelerate the self-exploration process. Based on this idea, we propose the extension of GAIC: GAIC$_{es}$, which can seamlessly integrate extra supervision into the self-exploration training pipeline. More specifically, we utilize the pixel-level person segmentation masks to guide the network to focus on the described person for gender identification.  Given the person segmentation masks, GAIC$_{es}$ has another loss, Gender Attention Loss $\mathcal{L}_{ga}$, to further fine-tune the learning process of $\alpha^g$. Note that the way to generate attention maps of gender words $\alpha^g$ in GAIC$_{es}$ is the same as that in GAIC. $\mathcal{L}_{ga}$ is defined as follows:
\begin{equation}
    \mathcal{L}_{ga} = 1 -  \sum\limits(\alpha^g\odot(1-M)),
\label{eq:mask}
\end{equation}
where $M$ denotes the binary pixel-level person segmentation mask which has values 1 in regions of person and 0 elsewhere. Loss $\mathcal{L}_{ga}$ encourages the attention maps of gender words not to exceed the regions of person and thus accelerates the learning process of gender features. The final loss of $\rm GAIC_{es}$ is defined as follows:
\begin{equation}
    \mathcal{L}_{es} = \mathcal{L}_{self} +
    \eta\mathcal{L}_{ga},
\end{equation}
where $\eta$ is the weighting parameter depending on how much emphasis we want to put on the extra supervision. Since labeling pixel-level segmentation maps are extremely time-consuming and costly, we prefer to use a very small amount of images with external supervision, e.g., 10\% in the following experiments. In Fig.~\ref{fig:model_pipeline}, we utilize $S_{es}$ to denote the extra supervision stream, and all three streams $S_{cg}$, $S_{gm}$ and $S_{es}$ share same parameters with each other and can be optimized in an end-to-end manner.

\begin{table*}[]
\centering
\begin{tabular}{cccccccccc}
\toprule
\multicolumn{1}{c|}{\multirow{2}{*}{Model}} &  \multicolumn{1}{c|}{\multirow{2}{*}{CIDEr}} & \multicolumn{1}{c|}{\multirow{2}{*}{METOR}} & \multicolumn{3}{c|}{Woman} & \multicolumn{3}{c|}{Men} &
\multicolumn{1}{c}{\multirow{2}{*}{Divergence}}\\
\cline{4-9} 
\multicolumn{1}{c|}{} & \multicolumn{1}{l|}{} & \multicolumn{1}{l|}{} & \multicolumn{1}{l|}{correct} & \multicolumn{1}{l|}{wrong} & \multicolumn{1}{l|}{neutral} & \multicolumn{1}{l|}{correct} & \multicolumn{1}{l|}{wrong} & 
\multicolumn{1}{l|}{neutral} \\ \hline
Att & 95.1 & 24.8 & 53.2 & 25.1 & 22.7 & 62.7 & 4.4 & 32.9 & 0.063\\
Balanced & 93.9 & 24.9 & 54.3 & 24.3 & 21.4 & 69.7 & 7.4 & 22.8 & 0.032\\
UpWeight-5 & 94.1 & 24.5 & 60.3 & 22.4 & 17.3 & 73.2 & 8.1 & 18.7 & 0.028\\
UpWeight-10 & 93.5 & 24.2 & \textbf{70.6} & 26.4 & 2.9 & \textbf{81.4} & 10.5 & 8.9 & 0.028\\
PixelSup & 92.0 & 24.5 & 57.3 & 21.1 & 21.6 & 68.3 & 8.6 & 23.1 & 0.032 \\
\hline
GAIC & 93.7 & 24.6 & 62.0 & 16.9 & 21.1 & 77.3 & 4.2 & 18.5  & 0.024  \\
GAIC$_{es}$ & 94.6 & 24.7 & 64.1 & \textbf{13.1} & 22.8 & 75.3 & \textbf{4.0} & 20.7 & \textbf{0.011}  \\
\bottomrule
\vspace{-10pt}
\end{tabular}

\caption{Gender bias analysis on COCO-GB V1 split. GAIC and GAIC$_{es}$ significantly improves the gender prediction performance compared to the baselines. Although UpWeight-10 model obtains the highest accuracy of both women and men, it also causes an unacceptable high error rate for two genders. For fairness evaluation, GAIC and GAIC$_{es}$ obtain the lowest divergence, which indicates that women and men have a similar outcome distribution and thus can be considered as more fair.}
\label{tab:COCO-GB V1}
\vspace{-15pt}
\end{table*}

\begin{table*}
\centering
\subfloat[Attention Sum]{%
\begin{tabular}{cccc}
\bottomrule
Accuracy & Women & Men & Average \\
\hline
Baseline & 25.5 & 21.2 & 23.4 \\
Balanced & 25.0 & 21.4 & 23.2 \\
Upweight-10 & 26.7 & 23.3 & 25.0 \\
PixelSup & 30.0 & 28.1 & 29.0 \\
\hline
GAIC & 27.4 & 24.3 & 25.6 \\
GAIC$_{es}$ & \textbf{32.5} & \textbf{28.5} & \textbf{30.1} \\
\bottomrule
\end{tabular}}%
\qquad
\subfloat[Point Game]{%
\begin{tabular}{cccc}
\bottomrule
Accuracy & Women & Men & Average\\
\hline
Baseline & 64.8 & 57.4 & 61.1 \\
Balanced & 66.2 & 59.6 & 62.9 \\
Upweight-10 & 66.2 & 59.3 & 62.8 \\
PixelSup & 67.2 & 60.5 & 63.9 \\
\hline
GAIC & 67.2 & 61.2 & 64.2 \\
GAIC$_{es}$ & \textbf{67.8} & \textbf{61.5} & \textbf{64.7} \\
\bottomrule
\end{tabular}}
\vspace{-10pt}
\caption{Attention Correctness on COCO-GB V1 split. We adopt Point Game \cite{zhang2018top} and Attention Sum \cite{liu2017attention} as evaluation metrics. Pointing Game measures the probability that point with highest attention value falls in the person segmentation masks. Attention Sum calculates the sum of attention weights that fall in the regions of person. The high outcome represents the correct of the model attention.  Results show that GAIC and GAIC$_{es}$ model can significantly improve attention correctness.}
\label{Tab: attention correctness}
\vspace{-20pt}
\end{table*}

\section{Experiments}
\subsection{Experiment Settings and Baselines} 
In our experiments, GAIC model is built on a widely used captioning model Att \cite{xu2015show} that is proved to learn gender bias in the benchmarking experiments. For $\rm GAIC_{es}$ model, we use 10\% images with person segmentation masks as the extra supervision. All models are trained with stochastic gradient descent using adaptive learning rate algorithms. We use early stopping on BLEU score to select the best model and set $\sigma=10, \mu=0.1, \eta=0.05$ in our experiments. 
We compare GAIC with the following three debiasing baselines:
\begin{itemize}[leftmargin=*]
\setlength\itemsep{0em}
\item \textbf{Balanced:} A subset is selected from the training data, which has a balanced gender ratio (4,000 images for each gender). We then fine-tune Att model on this new dataset. This baseline helps investigate the correlation between gender ratio and gender bias.

\vspace{2pt}
\item \textbf{UpWeight:} We also conduct an experiment with up-weighting the loss value of gender words' during training. To this end, we label the gender words position for each ground truth caption and multiply a constant value on the loss of gender words (we set the constant value to 5 and 10, which is denoted as UpWeight-5 and UpWeight-10). UpWeight forces the model to accurately predict gender words. However, unlike the Gender Evidence Loss, UpWeight does not encourage the model to predict gender neutral words when gender evidence is too vague to identify.

\vspace{2pt}
\item \textbf{Pixel-level Supervision (PixelSup):} As a variant of GAIC$_{es}$ model, we remove self-exploration streams and directly use Eq.~\ref{eq:mask} to fine-tune model attention with 10\% extra data, force gender words attention $\alpha_g$ not to exceed the person segmentation masks.  
\label{sec:baseline of debiasing}
\end{itemize}

\subsection{Evaluation Metrics} 
We consider the following metrics to evaluate captioning models:
\begin{itemize}[leftmargin=*]
\setlength\itemsep{0em}
\item
\textbf{Gender Accuracy:} Unlike the traditional binary gender classification, in this work, gender prediction results are grouped into correct, wrong, and neutral three categories. Due to the sensitive nature of prediction for protected attributes, reducing the error rate is most important. 
On this basis, the captioning model should predict gender cautiously and outputs neutral words when gender is too vague to distinguish.

\item\textbf{Gender Divergence:} We also expect models to treat each gender fairly. To this end, inspired by \cite{hendricks2018women}, we utilize Cosine Distance to measure the outcome similarity between Women and Men. For a fair system, it should have similar outcomes (correct/wrong/neutral rate) across different groups, which resembles to the fairness definition proposed in Equality of Odds \cite{hardt2016equality}. Lower divergence indicates that Women and Men have a similar distribution of outcomes and can be considered as more fair.

\item\textbf{Attention Correctness:} To measure whether attention focuses on the correct regions, we compare the attention maps of gender words with the person segmentation masks. We adopt two evaluation metrics to calculate the similarity: Pointing Game \cite{zhang2018top}, which measures whether the point with highest attention value falls in the person segmentation masks, and Attention Sum \cite{liu2017attention}, which calculates the sum of attention weights that fall in the person regions. For the two metrics, high outcome value represents the high similarity and thus can be consider as more correct.
\end{itemize}

\subsection{Experimental Results} 
\noindent\textbf{GAIC:} Tab.~2 reports the gender prediction results on COCO-GB V1. We observe that GAIC significantly improves the gender prediction performance compared to the base model Att. The gender accuracy of women increases from 53.2\% to 62.0\%, and its error rate reduces from 25.1\% to 16.9\%. The gender accuracy of men increases from 62.7\% to 77.3\%, and its error rate reduces from 4.4\% to 4.2\%.  

\vspace{1pt}
\noindent\textbf{UpWeight:} UpWeight-10 model obtains the highest accuracy of both women and men. However, it also causes the highest error rate for two genders, which is unacceptable. UpWeight-5 obtains a comparable gender correct rate compared with GAIC but has a notable higher error rate. The UpWeight model shows that up-weighting the loss value of gender words will proportionally increase both the correct and error rate and dramatically reduce the neutral rate. 

\vspace{1pt}
\noindent\textbf{Balanced:} An interesting finding is that there is no substantial difference between the Balanced model and Att model, a similar result is found by \cite{bolukbasi2016man}, which indicates that models learn gender bias mainly from bias in gender-context co-occurrence. Simply balancing the gender ratio of training samples thus has a limited improvement. A more efficient way is to eliminate gender-context bias. However, we emphasize that balancing the distribution for every gender-context pair in large-scale data is extremely difficult and guaranteeing complete gender-context decoupling is infeasible. 

\vspace{1pt}
\noindent\textbf{PixelSup:} We observe that PixelSup model obtains sensible improvements, which indicates that directly adding supervision on model's attention can control caption word prediction. 

\vspace{1pt}
\noindent\textbf{GAIC$_{es}$:} Compared to GAIC, GAIC$_{es}$ obtains consistently better performance. We empirically find that adding extra supervision accelerates the self-exploration process and makes training process more stable. To formalize the notion of fairness, we calculate the outcome divergence. GAIC and GAIC$_{es}$ have the lowest divergence, which indicates that women and men have a similar outcome distribution and thus can be considered as more fair.

Experiments on COCO-GB V2 have similar trends in COCO-GB V1 (detailed analysis of COCO-GB V2 results is in Sec.~\ref{sec:COCO-GB V2 result}). Compared to COCO-GB V1, all baselines in COCO-GB V2 obtain a worse gender prediction result. This is mainly because the unseen gender-context pairs in the test dataset increase the prediction difficulty. In comparison, our proposed GAIC and GAIC$_{es}$ model obtain a comparable performance on COCO-GB V1 dataset, which proves the robustness of the proposed self-exploration training strategy.

\begin{figure*}[t]
    \centering
    \includegraphics[width=0.86\linewidth]{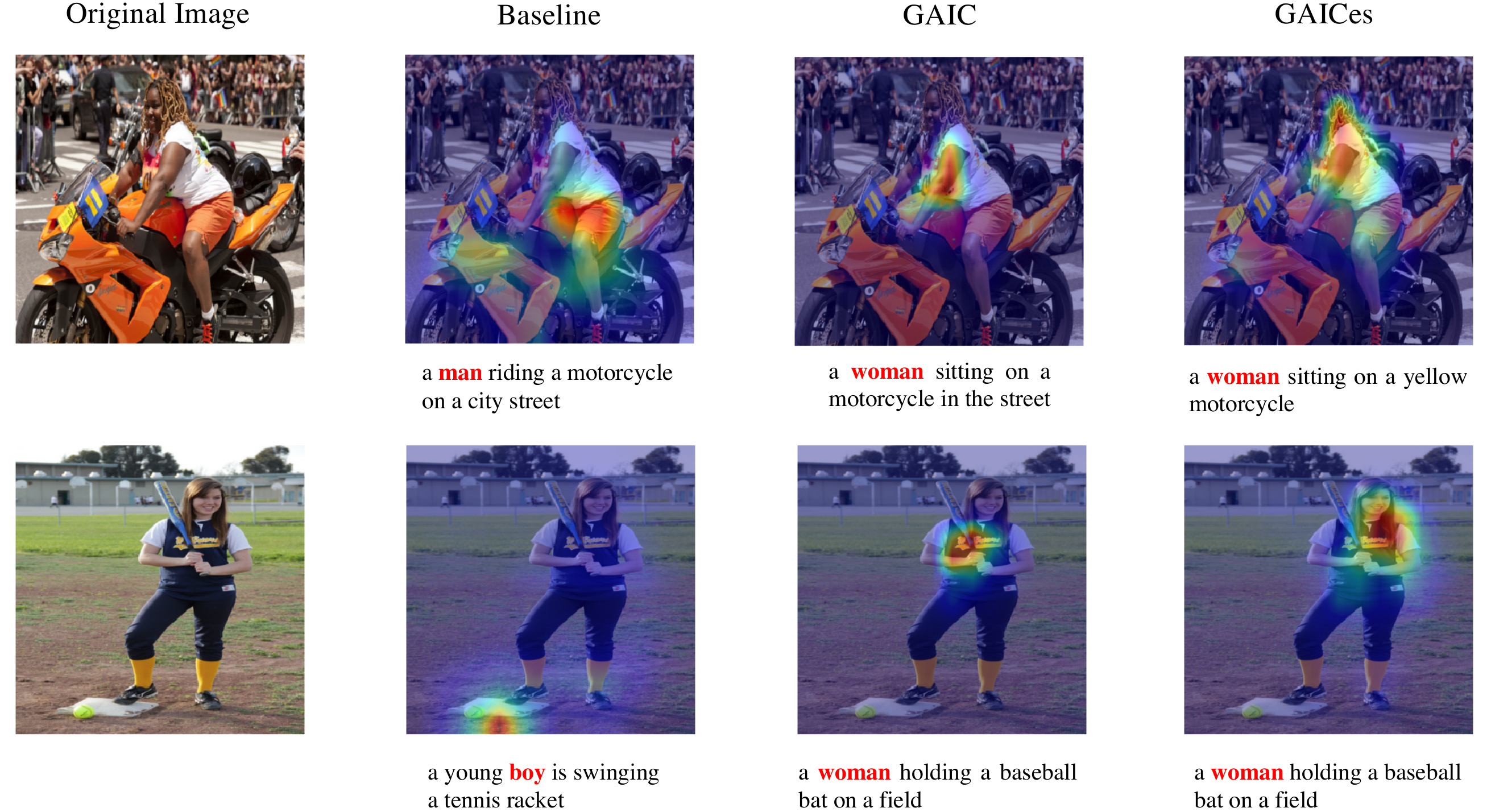}
    \vspace{-5pt}
    \caption{Qualitative comparison of baseline and our proposed models.} 
    \label{fig:attention_samples}
    \vspace{-10pt}
\end{figure*}

\vspace{2pt}
\noindent\textbf{Caption Quality:} Besides gender accuracy, we also expect the model to generate linguistically fluent captions. We use METEOR(M) and CIDEr(C) to evaluate caption quality. Results are shown in Tab.~2, GAIC and GAIC$_{es}$ only cause a minor performance drop compared to the baseline (drop from 95.1 to 94.6 on METROR and drop from 24.8 to 24.7 on CIDEr). In Fig.~\ref{fig:attention_samples}, examples show that the sentences generated by GAIC and GAIC$_{es}$ are linguistically fluent with more correct gender descriptions.

\vspace{2pt}
\noindent\textbf{Attention Correctness:} 
 To measure whether the model focuses on the correct gender features, we calculate the similarity between attention maps of gender words and the person segmentation masks. Quantitative results are shown in Tab.~\ref{Tab: attention correctness}. We observe that GAIC and GAIC$_{es}$ receive consistent improvement over the Att model and all variants on two evaluation metrics: Pointing Game and Attention Sum, which indicates that the proposed models tend to focus on the described person for the gender word prediction.  We also show qualitative comparisons in Fig.~\ref{fig:attention_samples}. We observe that the baseline Att model utilizes biased visual features and thus makes incorrect gender prediction, e.g., predicting a woman as a boy based on a tennis ball. In comparison, GAIC and GAIC$_{es}$ learn to concentrate on the regions of the described person for gender word prediction. In addition to correct gender prediction, proposed models learn to use gender neutral words when gender is too vague to distinguish. We put more visualizing attention results and discussions in Sec.~\ref{sec:More Qualitative Results}.

\section{Related Work}

\textbf{Gender Bias in Dataset:} The issue of gender bias has been studied in a variety of AI domains \cite{bender2018data, bolukbasi2016man, bordia2019identifying, brunet2019understanding, buolamwini2018gender, font2019equalizing, du2020fairness}, especially in natural language processing domain \cite{hendricks2018women, lambrecht2019algorithmic, rudinger2018gender, stock2018convnets, vanmassenhove, zhao2019gender, zhao2017men}. It has been reported that many popular language data resources contain gender bias \cite{wagner2015s, ross2011women, wiegand2019detection, sun2019mitigating}. As a result, the language models trained on the dataset may learn the bias and thus has preference or prejudice toward one gender over the other. Several studies have shown that the language models can learn and even amplify the gender bias in the dataset \cite{bolukbasi2016man, hendricks2018women, stock2018convnets, zhao2017men}. For instance, in visual semantic role labeling task, it shows that cooking images in the training set are twice that more likely involve females than males, and the model trained on the dataset further amplifies the disparity to five times during inference \cite{zhao2017men}. In regard to image captioning task, several initial attempts have been devoted to study gender bias in the captioning dataset. An early work analyzes the stereotype in Flicker30k dataset and mentions the gender-related issues \cite{mithun2018webly}. Another work demonstrates that caption annotators in COCO dataset tend to infer a person's gender from the image context when gender cannot be identified, e.g., annotators tend to describe snowboard players as “man” even gender is usually vague for the snowboard player \cite{hendricks2018women}.

\vspace{2pt}
\noindent\textbf{Vision and Language Bias:} Bias in captioning models can be grouped into vision and language bias \cite{rohrbach2018object}. The vision bias refers to the use of wrong visual evidence for caption generation. While the language bias refers to capture unwanted language priors in human annotations \cite{rohrbach2018object}. For example, the phrase "in a suit" often follows the word "man" in the training data. RNN's recurrent mechanism enables caption models to learn this language prior and predict the phrase "in a suit" after the word "man" regardless of the image contents. In this work, we notice that gender words are usually mentioned at the beginning of the sentence (on the average at position 2 with an average sentence length of 9), and the words inferred before gender words, such as "a" and "the," do not have any gender preference, which indicates that gender words cannot affect by the predicted words. Hence the gender bias problem we studied in this work should mainly come from the vision part. 


\noindent\textbf{Mitigating Gender Bias:} Few initial attempts have been made to  overcome gender biases in captioning models.
One solution is to make image captioning a two-step task \cite{bhargava2019exposing}. Firstly, an object detection network locates and recognizes the people in the image. Then a language model combines this extra grounded information with image embeddings to generate the final captions. For this method, gender prediction accuracy highly depends on the object detection model. Compared to the two-step training strategy, GAIC has an end-to-end architecture and thus does not rely on other grounded information. In another work \cite{hendricks2018women}, the authors alleviate the gender bias by modifying the training images 
and loss functions. However, their approach requires a high-quality person segmentation mask for each training data, which is costly and infeasible for many large-scale datasets. In comparison, the proposed self-guided attention mechanism can work without extra supervision and thus greatly expands the application scenarios.

\section{Conclusions and Future Work} In this paper, two novel COCO splits are created for studying gender bias problem in image captioning task. We provide extensive baseline experiments for benchmarking different models, training strategies, as well as a comprehensive analysis of our new datasets. The experimental results indicate that many captioning models utilize contextual cues for gender identification, leading to undesirable gender prediction errors, especially for women. To overcome this unwanted bias, we propose a new training framework GAIC, which can significantly reduce gender bias by self-guided supervision. Experimental results validate that the proposed models learn to focus on the described person for gender identification.

As an initial attempt to approach the potential gender bias in captioning systems, we emphasize that utilizing the gender prediction accuracy to quantify gender bias is not enough, evaluation metrics that can effectively lead to high caption quality and low gender bias are still lacking. Understanding and evaluating the caption bias from the causal inference perspective is a promising research field and would be exploded in our future research.

\begin{acks}
The authors thank the anonymous reviewers for their helpful comments. The work is in part supported by NSF IIS-1900990, and NSF IIS-1939716. The views and conclusions contained in this paper are those of the authors and should not be interpreted as representing any funding agencies.
\end{acks}

\newpage
\bibliographystyle{ACM-Reference-Format}
\bibliography{ref}


\begin{thebibliography}{46}


\ifx \showCODEN    \undefined \def \showCODEN     #1{\unskip}     \fi
\ifx \showDOI      \undefined \def \showDOI       #1{#1}\fi
\ifx \showISBNx    \undefined \def \showISBNx     #1{\unskip}     \fi
\ifx \showISBNxiii \undefined \def \showISBNxiii  #1{\unskip}     \fi
\ifx \showISSN     \undefined \def \showISSN      #1{\unskip}     \fi
\ifx \showLCCN     \undefined \def \showLCCN      #1{\unskip}     \fi
\ifx \shownote     \undefined \def \shownote      #1{#1}          \fi
\ifx \showarticletitle \undefined \def \showarticletitle #1{#1}   \fi
\ifx \showURL      \undefined \def \showURL       {\relax}        \fi
\providecommand\bibfield[2]{#2}
\providecommand\bibinfo[2]{#2}
\providecommand\natexlab[1]{#1}
\providecommand\showeprint[2][]{arXiv:#2}

\bibitem[\protect\citeauthoryear{Agrawal, Batra, Parikh, and Kembhavi}{Agrawal
  et~al\mbox{.}}{2018}]%
        {agrawal2018don}
\bibfield{author}{\bibinfo{person}{Aishwarya Agrawal}, \bibinfo{person}{Dhruv
  Batra}, \bibinfo{person}{Devi Parikh}, {and} \bibinfo{person}{Aniruddha
  Kembhavi}.} \bibinfo{year}{2018}\natexlab{}.
\newblock \showarticletitle{Don't just assume; look and answer: Overcoming
  priors for visual question answering}. In
  \bibinfo{booktitle}{\emph{Proceedings of the IEEE Conference on Computer
  Vision and Pattern Recognition}}. \bibinfo{pages}{4971--4980}.
\newblock


\bibitem[\protect\citeauthoryear{Anderson, He, Buehler, Teney, Johnson, Gould,
  and Zhang}{Anderson et~al\mbox{.}}{2018}]%
        {anderson2018bottom}
\bibfield{author}{\bibinfo{person}{Peter Anderson}, \bibinfo{person}{Xiaodong
  He}, \bibinfo{person}{Chris Buehler}, \bibinfo{person}{Damien Teney},
  \bibinfo{person}{Mark Johnson}, \bibinfo{person}{Stephen Gould}, {and}
  \bibinfo{person}{Lei Zhang}.} \bibinfo{year}{2018}\natexlab{}.
\newblock \showarticletitle{Bottom-Up and Top-Down Attention for Image
  Captioning and Visual Question Answering}. In \bibinfo{booktitle}{\emph{2018
  IEEE/CVF Conference on Computer Vision and Pattern Recognition (CVPR)}}.
  IEEE, \bibinfo{pages}{6077--6086}.
\newblock


\bibitem[\protect\citeauthoryear{Bender and Friedman}{Bender and
  Friedman}{2018}]%
        {bender2018data}
\bibfield{author}{\bibinfo{person}{Emily~M Bender} {and} \bibinfo{person}{Batya
  Friedman}.} \bibinfo{year}{2018}\natexlab{}.
\newblock \showarticletitle{Data Statements for Natural Language Processing:
  Toward Mitigating System Bias and Enabling Better Science}.
\newblock \bibinfo{journal}{\emph{Transactions of the Association for
  Computational Linguistics}}  \bibinfo{volume}{6} (\bibinfo{year}{2018}),
  \bibinfo{pages}{587--604}.
\newblock


\bibitem[\protect\citeauthoryear{Bhargava}{Bhargava}{2019}]%
        {bhargava2019exposing}
\bibfield{author}{\bibinfo{person}{Shruti Bhargava}.}
  \bibinfo{year}{2019}\natexlab{}.
\newblock \emph{\bibinfo{title}{Exposing and correcting the gender bias in
  image captioning datasets and models}}.
\newblock \bibinfo{thesistype}{Ph.D. Dissertation}.
\newblock


\bibitem[\protect\citeauthoryear{Bolukbasi, Chang, Zou, Saligrama, and
  Kalai}{Bolukbasi et~al\mbox{.}}{2016}]%
        {bolukbasi2016man}
\bibfield{author}{\bibinfo{person}{Tolga Bolukbasi}, \bibinfo{person}{Kai-Wei
  Chang}, \bibinfo{person}{James~Y Zou}, \bibinfo{person}{Venkatesh Saligrama},
  {and} \bibinfo{person}{Adam~T Kalai}.} \bibinfo{year}{2016}\natexlab{}.
\newblock \showarticletitle{Man is to computer programmer as woman is to
  homemaker? debiasing word embeddings}. In \bibinfo{booktitle}{\emph{Advances
  in neural information processing systems}}. \bibinfo{pages}{4349--4357}.
\newblock


\bibitem[\protect\citeauthoryear{Bordia and Bowman}{Bordia and Bowman}{2019}]%
        {bordia2019identifying}
\bibfield{author}{\bibinfo{person}{Shikha Bordia} {and} \bibinfo{person}{Samuel
  Bowman}.} \bibinfo{year}{2019}\natexlab{}.
\newblock \showarticletitle{Identifying and Reducing Gender Bias in Word-Level
  Language Models}. In \bibinfo{booktitle}{\emph{Proceedings of the 2019
  Conference of the North American Chapter of the Association for Computational
  Linguistics: Student Research Workshop}}. \bibinfo{pages}{7--15}.
\newblock


\bibitem[\protect\citeauthoryear{Brunet, Alkalay-Houlihan, Anderson, and
  Zemel}{Brunet et~al\mbox{.}}{2019}]%
        {brunet2019understanding}
\bibfield{author}{\bibinfo{person}{Marc-Etienne Brunet},
  \bibinfo{person}{Colleen Alkalay-Houlihan}, \bibinfo{person}{Ashton
  Anderson}, {and} \bibinfo{person}{Richard Zemel}.}
  \bibinfo{year}{2019}\natexlab{}.
\newblock \showarticletitle{Understanding the Origins of Bias in Word
  Embeddings}. In \bibinfo{booktitle}{\emph{International Conference on Machine
  Learning}}. \bibinfo{pages}{803--811}.
\newblock


\bibitem[\protect\citeauthoryear{Buolamwini and Gebru}{Buolamwini and
  Gebru}{2018}]%
        {buolamwini2018gender}
\bibfield{author}{\bibinfo{person}{Joy Buolamwini} {and}
  \bibinfo{person}{Timnit Gebru}.} \bibinfo{year}{2018}\natexlab{}.
\newblock \showarticletitle{Gender shades: Intersectional accuracy disparities
  in commercial gender classification}. In \bibinfo{booktitle}{\emph{Conference
  on fairness, accountability and transparency}}. \bibinfo{pages}{77--91}.
\newblock


\bibitem[\protect\citeauthoryear{Denkowski and Lavie}{Denkowski and
  Lavie}{2011}]%
        {denkowski2011meteor}
\bibfield{author}{\bibinfo{person}{Michael Denkowski} {and}
  \bibinfo{person}{Alon Lavie}.} \bibinfo{year}{2011}\natexlab{}.
\newblock \showarticletitle{Meteor 1.3: Automatic metric for reliable
  optimization and evaluation of machine translation systems}. In
  \bibinfo{booktitle}{\emph{Proceedings of the sixth workshop on statistical
  machine translation}}. \bibinfo{pages}{85--91}.
\newblock


\bibitem[\protect\citeauthoryear{Donahue, Anne~Hendricks, Guadarrama, Rohrbach,
  Venugopalan, Saenko, and Darrell}{Donahue et~al\mbox{.}}{2015}]%
        {donahue2015long}
\bibfield{author}{\bibinfo{person}{Jeffrey Donahue}, \bibinfo{person}{Lisa
  Anne~Hendricks}, \bibinfo{person}{Sergio Guadarrama}, \bibinfo{person}{Marcus
  Rohrbach}, \bibinfo{person}{Subhashini Venugopalan}, \bibinfo{person}{Kate
  Saenko}, {and} \bibinfo{person}{Trevor Darrell}.}
  \bibinfo{year}{2015}\natexlab{}.
\newblock \showarticletitle{Long-term recurrent convolutional networks for
  visual recognition and description}. In \bibinfo{booktitle}{\emph{Proceedings
  of the IEEE conference on computer vision and pattern recognition}}.
  \bibinfo{pages}{2625--2634}.
\newblock


\bibitem[\protect\citeauthoryear{Du, Yang, Zou, and Hu}{Du
  et~al\mbox{.}}{2020}]%
        {du2020fairness}
\bibfield{author}{\bibinfo{person}{Mengnan Du}, \bibinfo{person}{Fan Yang},
  \bibinfo{person}{Na Zou}, {and} \bibinfo{person}{Xia Hu}.}
  \bibinfo{year}{2020}\natexlab{}.
\newblock \showarticletitle{Fairness in deep learning: A computational
  perspective}.
\newblock \bibinfo{journal}{\emph{IEEE Intelligent Systems}}
  (\bibinfo{year}{2020}).
\newblock


\bibitem[\protect\citeauthoryear{Fang, Gupta, Iandola, Srivastava, Deng,
  Doll{\'a}r, Gao, He, Mitchell, Platt, et~al\mbox{.}}{Fang
  et~al\mbox{.}}{2015}]%
        {fang2015captions}
\bibfield{author}{\bibinfo{person}{Hao Fang}, \bibinfo{person}{Saurabh Gupta},
  \bibinfo{person}{Forrest Iandola}, \bibinfo{person}{Rupesh~K Srivastava},
  \bibinfo{person}{Li Deng}, \bibinfo{person}{Piotr Doll{\'a}r},
  \bibinfo{person}{Jianfeng Gao}, \bibinfo{person}{Xiaodong He},
  \bibinfo{person}{Margaret Mitchell}, \bibinfo{person}{John~C Platt},
  {et~al\mbox{.}}} \bibinfo{year}{2015}\natexlab{}.
\newblock \showarticletitle{From captions to visual concepts and back}. In
  \bibinfo{booktitle}{\emph{Proceedings of the IEEE conference on computer
  vision and pattern recognition}}. \bibinfo{pages}{1473--1482}.
\newblock


\bibitem[\protect\citeauthoryear{Font and Costa-juss{\`a}}{Font and
  Costa-juss{\`a}}{2019}]%
        {font2019equalizing}
\bibfield{author}{\bibinfo{person}{Joel~Escud{\'e} Font} {and}
  \bibinfo{person}{Marta~R Costa-juss{\`a}}.} \bibinfo{year}{2019}\natexlab{}.
\newblock \showarticletitle{Equalizing Gender Bias in Neural Machine
  Translation with Word Embeddings Techniques}. In
  \bibinfo{booktitle}{\emph{Proceedings of the First Workshop on Gender Bias in
  Natural Language Processing}}. \bibinfo{pages}{147--154}.
\newblock


\bibitem[\protect\citeauthoryear{Geirhos, Jacobsen, Michaelis, Zemel, Brendel,
  Bethge, and Wichmann}{Geirhos et~al\mbox{.}}{2020}]%
        {geirhos2020shortcut}
\bibfield{author}{\bibinfo{person}{Robert Geirhos},
  \bibinfo{person}{J{\"o}rn-Henrik Jacobsen}, \bibinfo{person}{Claudio
  Michaelis}, \bibinfo{person}{Richard Zemel}, \bibinfo{person}{Wieland
  Brendel}, \bibinfo{person}{Matthias Bethge}, {and} \bibinfo{person}{Felix~A
  Wichmann}.} \bibinfo{year}{2020}\natexlab{}.
\newblock \showarticletitle{Shortcut Learning in Deep Neural Networks}.
\newblock \bibinfo{journal}{\emph{arXiv preprint arXiv:2004.07780}}
  (\bibinfo{year}{2020}).
\newblock


\bibitem[\protect\citeauthoryear{Hardt, Price, and Srebro}{Hardt
  et~al\mbox{.}}{2016}]%
        {hardt2016equality}
\bibfield{author}{\bibinfo{person}{Moritz Hardt}, \bibinfo{person}{Eric Price},
  {and} \bibinfo{person}{Nati Srebro}.} \bibinfo{year}{2016}\natexlab{}.
\newblock \showarticletitle{Equality of opportunity in supervised learning}. In
  \bibinfo{booktitle}{\emph{Advances in neural information processing
  systems}}. \bibinfo{pages}{3315--3323}.
\newblock


\bibitem[\protect\citeauthoryear{Hendricks, Burns, Saenko, Darrell, and
  Rohrbach}{Hendricks et~al\mbox{.}}{2018}]%
        {hendricks2018women}
\bibfield{author}{\bibinfo{person}{Lisa~Anne Hendricks},
  \bibinfo{person}{Kaylee Burns}, \bibinfo{person}{Kate Saenko},
  \bibinfo{person}{Trevor Darrell}, {and} \bibinfo{person}{Anna Rohrbach}.}
  \bibinfo{year}{2018}\natexlab{}.
\newblock \showarticletitle{Women also snowboard: Overcoming bias in captioning
  models}. In \bibinfo{booktitle}{\emph{European Conference on Computer
  Vision}}. Springer, \bibinfo{pages}{793--811}.
\newblock


\bibitem[\protect\citeauthoryear{Hochreiter and Schmidhuber}{Hochreiter and
  Schmidhuber}{1997}]%
        {hochreiter1997long}
\bibfield{author}{\bibinfo{person}{Sepp Hochreiter} {and}
  \bibinfo{person}{J{\"u}rgen Schmidhuber}.} \bibinfo{year}{1997}\natexlab{}.
\newblock \showarticletitle{Long short-term memory}.
\newblock \bibinfo{journal}{\emph{Neural computation}} \bibinfo{volume}{9},
  \bibinfo{number}{8} (\bibinfo{year}{1997}), \bibinfo{pages}{1735--1780}.
\newblock


\bibitem[\protect\citeauthoryear{Hossain, Sohel, Shiratuddin, and Laga}{Hossain
  et~al\mbox{.}}{2019}]%
        {hossain2019comprehensive}
\bibfield{author}{\bibinfo{person}{MD Hossain}, \bibinfo{person}{Ferdous
  Sohel}, \bibinfo{person}{Mohd~Fairuz Shiratuddin}, {and}
  \bibinfo{person}{Hamid Laga}.} \bibinfo{year}{2019}\natexlab{}.
\newblock \showarticletitle{A Comprehensive Survey of Deep Learning for Image
  Captioning}.
\newblock \bibinfo{journal}{\emph{ACM Computing Surveys (CSUR)}}
  \bibinfo{volume}{51}, \bibinfo{number}{6} (\bibinfo{year}{2019}),
  \bibinfo{pages}{118}.
\newblock


\bibitem[\protect\citeauthoryear{Karpathy and Fei-Fei}{Karpathy and
  Fei-Fei}{2015}]%
        {karpathy2015deep}
\bibfield{author}{\bibinfo{person}{Andrej Karpathy} {and} \bibinfo{person}{Li
  Fei-Fei}.} \bibinfo{year}{2015}\natexlab{}.
\newblock \showarticletitle{Deep visual-semantic alignments for generating
  image descriptions}. In \bibinfo{booktitle}{\emph{Proceedings of the IEEE
  conference on computer vision and pattern recognition}}.
  \bibinfo{pages}{3128--3137}.
\newblock


\bibitem[\protect\citeauthoryear{Lambrecht and Tucker}{Lambrecht and
  Tucker}{2019}]%
        {lambrecht2019algorithmic}
\bibfield{author}{\bibinfo{person}{Anja Lambrecht} {and}
  \bibinfo{person}{Catherine Tucker}.} \bibinfo{year}{2019}\natexlab{}.
\newblock \showarticletitle{Algorithmic bias? an empirical study of apparent
  gender-based discrimination in the display of stem career ads}.
\newblock \bibinfo{journal}{\emph{Management Science}} \bibinfo{volume}{65},
  \bibinfo{number}{7} (\bibinfo{year}{2019}), \bibinfo{pages}{2966--2981}.
\newblock


\bibitem[\protect\citeauthoryear{Li, Wu, Peng, Ernst, and Fu}{Li
  et~al\mbox{.}}{2018}]%
        {li2018tell}
\bibfield{author}{\bibinfo{person}{Kunpeng Li}, \bibinfo{person}{Ziyan Wu},
  \bibinfo{person}{Kuan-Chuan Peng}, \bibinfo{person}{Jan Ernst}, {and}
  \bibinfo{person}{Yun Fu}.} \bibinfo{year}{2018}\natexlab{}.
\newblock \showarticletitle{Tell me where to look: Guided attention inference
  network}. In \bibinfo{booktitle}{\emph{Proceedings of the IEEE Conference on
  Computer Vision and Pattern Recognition}}. \bibinfo{pages}{9215--9223}.
\newblock


\bibitem[\protect\citeauthoryear{Lin, Maire, Belongie, Hays, Perona, Ramanan,
  Doll{\'a}r, and Zitnick}{Lin et~al\mbox{.}}{2014}]%
        {lin2014microsoft}
\bibfield{author}{\bibinfo{person}{Tsung-Yi Lin}, \bibinfo{person}{Michael
  Maire}, \bibinfo{person}{Serge Belongie}, \bibinfo{person}{James Hays},
  \bibinfo{person}{Pietro Perona}, \bibinfo{person}{Deva Ramanan},
  \bibinfo{person}{Piotr Doll{\'a}r}, {and} \bibinfo{person}{C~Lawrence
  Zitnick}.} \bibinfo{year}{2014}\natexlab{}.
\newblock \showarticletitle{Microsoft coco: Common objects in context}. In
  \bibinfo{booktitle}{\emph{European conference on computer vision}}. Springer,
  \bibinfo{pages}{740--755}.
\newblock


\bibitem[\protect\citeauthoryear{Liu, Mao, Sha, and Yuille}{Liu
  et~al\mbox{.}}{2017}]%
        {liu2017attention}
\bibfield{author}{\bibinfo{person}{Chenxi Liu}, \bibinfo{person}{Junhua Mao},
  \bibinfo{person}{Fei Sha}, {and} \bibinfo{person}{Alan Yuille}.}
  \bibinfo{year}{2017}\natexlab{}.
\newblock \showarticletitle{Attention correctness in neural image captioning}.
  In \bibinfo{booktitle}{\emph{Thirty-First AAAI Conference on Artificial
  Intelligence}}.
\newblock


\bibitem[\protect\citeauthoryear{Lu, Xiong, Parikh, and Socher}{Lu
  et~al\mbox{.}}{2017}]%
        {lu2017knowing}
\bibfield{author}{\bibinfo{person}{Jiasen Lu}, \bibinfo{person}{Caiming Xiong},
  \bibinfo{person}{Devi Parikh}, {and} \bibinfo{person}{Richard Socher}.}
  \bibinfo{year}{2017}\natexlab{}.
\newblock \showarticletitle{Knowing when to look: Adaptive attention via a
  visual sentinel for image captioning}. In
  \bibinfo{booktitle}{\emph{Proceedings of the IEEE conference on computer
  vision and pattern recognition}}. \bibinfo{pages}{375--383}.
\newblock


\bibitem[\protect\citeauthoryear{Lu, Yang, Batra, and Parikh}{Lu
  et~al\mbox{.}}{2018}]%
        {lu2018neural}
\bibfield{author}{\bibinfo{person}{Jiasen Lu}, \bibinfo{person}{Jianwei Yang},
  \bibinfo{person}{Dhruv Batra}, {and} \bibinfo{person}{Devi Parikh}.}
  \bibinfo{year}{2018}\natexlab{}.
\newblock \showarticletitle{Neural baby talk}. In
  \bibinfo{booktitle}{\emph{Proceedings of the IEEE conference on computer
  vision and pattern recognition}}. \bibinfo{pages}{7219--7228}.
\newblock


\bibitem[\protect\citeauthoryear{Mao, Xu, Yang, Wang, Huang, and Yuille}{Mao
  et~al\mbox{.}}{2014}]%
        {mao2014deep}
\bibfield{author}{\bibinfo{person}{Junhua Mao}, \bibinfo{person}{Wei Xu},
  \bibinfo{person}{Yi Yang}, \bibinfo{person}{Jiang Wang},
  \bibinfo{person}{Zhiheng Huang}, {and} \bibinfo{person}{Alan Yuille}.}
  \bibinfo{year}{2014}\natexlab{}.
\newblock \showarticletitle{Deep captioning with multimodal recurrent neural
  networks (m-rnn)}.
\newblock \bibinfo{journal}{\emph{arXiv preprint arXiv:1412.6632}}
  (\bibinfo{year}{2014}).
\newblock


\bibitem[\protect\citeauthoryear{Mithun, Panda, Papalexakis, and
  Roy-Chowdhury}{Mithun et~al\mbox{.}}{2018}]%
        {mithun2018webly}
\bibfield{author}{\bibinfo{person}{Niluthpol~Chowdhury Mithun},
  \bibinfo{person}{Rameswar Panda}, \bibinfo{person}{Evangelos~E Papalexakis},
  {and} \bibinfo{person}{Amit~K Roy-Chowdhury}.}
  \bibinfo{year}{2018}\natexlab{}.
\newblock \showarticletitle{Webly supervised joint embedding for cross-modal
  image-text retrieval}. In \bibinfo{booktitle}{\emph{Proceedings of the 26th
  ACM international conference on Multimedia}}. \bibinfo{pages}{1856--1864}.
\newblock


\bibitem[\protect\citeauthoryear{Papineni, Roukos, Ward, and Zhu}{Papineni
  et~al\mbox{.}}{2002}]%
        {papineni2002bleu}
\bibfield{author}{\bibinfo{person}{Kishore Papineni}, \bibinfo{person}{Salim
  Roukos}, \bibinfo{person}{Todd Ward}, {and} \bibinfo{person}{Wei-Jing Zhu}.}
  \bibinfo{year}{2002}\natexlab{}.
\newblock \showarticletitle{BLEU: a method for automatic evaluation of machine
  translation}. In \bibinfo{booktitle}{\emph{Proceedings of the 40th annual
  meeting on association for computational linguistics}}. Association for
  Computational Linguistics, \bibinfo{pages}{311--318}.
\newblock


\bibitem[\protect\citeauthoryear{Ren, He, Girshick, and Sun}{Ren
  et~al\mbox{.}}{2015}]%
        {ren2015faster}
\bibfield{author}{\bibinfo{person}{Shaoqing Ren}, \bibinfo{person}{Kaiming He},
  \bibinfo{person}{Ross Girshick}, {and} \bibinfo{person}{Jian Sun}.}
  \bibinfo{year}{2015}\natexlab{}.
\newblock \showarticletitle{Faster r-cnn: Towards real-time object detection
  with region proposal networks}. In \bibinfo{booktitle}{\emph{Advances in
  neural information processing systems}}. \bibinfo{pages}{91--99}.
\newblock


\bibitem[\protect\citeauthoryear{Rennie, Marcheret, Mroueh, Ross, and
  Goel}{Rennie et~al\mbox{.}}{2017}]%
        {rennie2017self}
\bibfield{author}{\bibinfo{person}{Steven~J Rennie}, \bibinfo{person}{Etienne
  Marcheret}, \bibinfo{person}{Youssef Mroueh}, \bibinfo{person}{Jerret Ross},
  {and} \bibinfo{person}{Vaibhava Goel}.} \bibinfo{year}{2017}\natexlab{}.
\newblock \showarticletitle{Self-critical sequence training for image
  captioning}. In \bibinfo{booktitle}{\emph{Proceedings of the IEEE Conference
  on Computer Vision and Pattern Recognition}}. \bibinfo{pages}{7008--7024}.
\newblock


\bibitem[\protect\citeauthoryear{Rohrbach, Hendricks, Burns, Darrell, and
  Saenko}{Rohrbach et~al\mbox{.}}{2018}]%
        {rohrbach2018object}
\bibfield{author}{\bibinfo{person}{Anna Rohrbach}, \bibinfo{person}{Lisa~Anne
  Hendricks}, \bibinfo{person}{Kaylee Burns}, \bibinfo{person}{Trevor Darrell},
  {and} \bibinfo{person}{Kate Saenko}.} \bibinfo{year}{2018}\natexlab{}.
\newblock \showarticletitle{Object Hallucination in Image Captioning}. In
  \bibinfo{booktitle}{\emph{Proceedings of the 2018 Conference on Empirical
  Methods in Natural Language Processing}}. \bibinfo{pages}{4035--4045}.
\newblock


\bibitem[\protect\citeauthoryear{Ross and Carter}{Ross and Carter}{2011}]%
        {ross2011women}
\bibfield{author}{\bibinfo{person}{Karen Ross} {and} \bibinfo{person}{Cynthia
  Carter}.} \bibinfo{year}{2011}\natexlab{}.
\newblock \showarticletitle{Women and news: A long and winding road}.
\newblock \bibinfo{journal}{\emph{Media, Culture \& Society}}
  \bibinfo{volume}{33}, \bibinfo{number}{8} (\bibinfo{year}{2011}),
  \bibinfo{pages}{1148--1165}.
\newblock


\bibitem[\protect\citeauthoryear{Rudinger, Naradowsky, Leonard, and
  Van~Durme}{Rudinger et~al\mbox{.}}{2018}]%
        {rudinger2018gender}
\bibfield{author}{\bibinfo{person}{Rachel Rudinger}, \bibinfo{person}{Jason
  Naradowsky}, \bibinfo{person}{Brian Leonard}, {and} \bibinfo{person}{Benjamin
  Van~Durme}.} \bibinfo{year}{2018}\natexlab{}.
\newblock \showarticletitle{Gender Bias in Coreference Resolution}. In
  \bibinfo{booktitle}{\emph{Proceedings of the 2018 Conference of the North
  American Chapter of the Association for Computational Linguistics: Human
  Language Technologies, Volume 2 (Short Papers)}}. \bibinfo{pages}{8--14}.
\newblock


\bibitem[\protect\citeauthoryear{Selvaraju, Cogswell, Das, Vedantam, Parikh,
  and Batra}{Selvaraju et~al\mbox{.}}{2017}]%
        {selvaraju2017grad}
\bibfield{author}{\bibinfo{person}{Ramprasaath~R Selvaraju},
  \bibinfo{person}{Michael Cogswell}, \bibinfo{person}{Abhishek Das},
  \bibinfo{person}{Ramakrishna Vedantam}, \bibinfo{person}{Devi Parikh}, {and}
  \bibinfo{person}{Dhruv Batra}.} \bibinfo{year}{2017}\natexlab{}.
\newblock \showarticletitle{Grad-cam: Visual explanations from deep networks
  via gradient-based localization}. In \bibinfo{booktitle}{\emph{Proceedings of
  the IEEE international conference on computer vision}}.
  \bibinfo{pages}{618--626}.
\newblock


\bibitem[\protect\citeauthoryear{Simonyan, Vedaldi, and Zisserman}{Simonyan
  et~al\mbox{.}}{2014}]%
        {simonyan2014deep}
\bibfield{author}{\bibinfo{person}{K Simonyan}, \bibinfo{person}{A Vedaldi},
  {and} \bibinfo{person}{A Zisserman}.} \bibinfo{year}{2014}\natexlab{}.
\newblock \showarticletitle{Deep inside convolutional networks: visualising
  image classification models and saliency maps}.
\newblock  (\bibinfo{year}{2014}).
\newblock


\bibitem[\protect\citeauthoryear{Stock and Cisse}{Stock and Cisse}{2018}]%
        {stock2018convnets}
\bibfield{author}{\bibinfo{person}{Pierre Stock} {and}
  \bibinfo{person}{Moustapha Cisse}.} \bibinfo{year}{2018}\natexlab{}.
\newblock \showarticletitle{Convnets and imagenet beyond accuracy:
  Understanding mistakes and uncovering biases}. In
  \bibinfo{booktitle}{\emph{Proceedings of the European Conference on Computer
  Vision (ECCV)}}. \bibinfo{pages}{498--512}.
\newblock


\bibitem[\protect\citeauthoryear{Sun, Gaut, Tang, Huang, ElSherief, Zhao,
  Mirza, Belding, Chang, and Wang}{Sun et~al\mbox{.}}{2019}]%
        {sun2019mitigating}
\bibfield{author}{\bibinfo{person}{Tony Sun}, \bibinfo{person}{Andrew Gaut},
  \bibinfo{person}{Shirlyn Tang}, \bibinfo{person}{Yuxin Huang},
  \bibinfo{person}{Mai ElSherief}, \bibinfo{person}{Jieyu Zhao},
  \bibinfo{person}{Diba Mirza}, \bibinfo{person}{Elizabeth Belding},
  \bibinfo{person}{Kai-Wei Chang}, {and} \bibinfo{person}{William~Yang Wang}.}
  \bibinfo{year}{2019}\natexlab{}.
\newblock \showarticletitle{Mitigating gender bias in natural language
  processing: Literature review}.
\newblock \bibinfo{journal}{\emph{arXiv preprint arXiv:1906.08976}}
  (\bibinfo{year}{2019}).
\newblock


\bibitem[\protect\citeauthoryear{Vanmassenhove, Hardmeier, and
  Way}{Vanmassenhove et~al\mbox{.}}{2018}]%
        {vanmassenhove}
\bibfield{author}{\bibinfo{person}{Eva Vanmassenhove},
  \bibinfo{person}{Christian Hardmeier}, {and} \bibinfo{person}{Andy Way}.}
  \bibinfo{year}{2018}\natexlab{}.
\newblock \showarticletitle{Getting Gender Right in Neural Machine
  Translation}. In \bibinfo{booktitle}{\emph{Proceedings of the 2018 Conference
  on Empirical Methods in Natural Language Processing}}.
  \bibinfo{pages}{3003--3008}.
\newblock


\bibitem[\protect\citeauthoryear{Vedantam, Lawrence~Zitnick, and
  Parikh}{Vedantam et~al\mbox{.}}{2015}]%
        {vedantam2015cider}
\bibfield{author}{\bibinfo{person}{Ramakrishna Vedantam}, \bibinfo{person}{C
  Lawrence~Zitnick}, {and} \bibinfo{person}{Devi Parikh}.}
  \bibinfo{year}{2015}\natexlab{}.
\newblock \showarticletitle{Cider: Consensus-based image description
  evaluation}. In \bibinfo{booktitle}{\emph{Proceedings of the IEEE conference
  on computer vision and pattern recognition}}. \bibinfo{pages}{4566--4575}.
\newblock


\bibitem[\protect\citeauthoryear{Vinyals, Toshev, Bengio, and Erhan}{Vinyals
  et~al\mbox{.}}{2015}]%
        {vinyals2015show}
\bibfield{author}{\bibinfo{person}{Oriol Vinyals}, \bibinfo{person}{Alexander
  Toshev}, \bibinfo{person}{Samy Bengio}, {and} \bibinfo{person}{Dumitru
  Erhan}.} \bibinfo{year}{2015}\natexlab{}.
\newblock \showarticletitle{Show and tell: A neural image caption generator}.
  In \bibinfo{booktitle}{\emph{Proceedings of the IEEE conference on computer
  vision and pattern recognition}}. \bibinfo{pages}{3156--3164}.
\newblock


\bibitem[\protect\citeauthoryear{Wagner, Garcia, Jadidi, and Strohmaier}{Wagner
  et~al\mbox{.}}{2015}]%
        {wagner2015s}
\bibfield{author}{\bibinfo{person}{Claudia Wagner}, \bibinfo{person}{David
  Garcia}, \bibinfo{person}{Mohsen Jadidi}, {and} \bibinfo{person}{Markus
  Strohmaier}.} \bibinfo{year}{2015}\natexlab{}.
\newblock \showarticletitle{It's a Man's Wikipedia? Assessing Gender Inequality
  in an Online Encyclopedia}. In \bibinfo{booktitle}{\emph{International AAAI
  Conference on Weblogs and Social Media}}. USA, \bibinfo{pages}{454--463}.
\newblock


\bibitem[\protect\citeauthoryear{Wiegand, Ruppenhofer, and Kleinbauer}{Wiegand
  et~al\mbox{.}}{2019}]%
        {wiegand2019detection}
\bibfield{author}{\bibinfo{person}{Michael Wiegand}, \bibinfo{person}{Josef
  Ruppenhofer}, {and} \bibinfo{person}{Thomas Kleinbauer}.}
  \bibinfo{year}{2019}\natexlab{}.
\newblock \showarticletitle{Detection of abusive language: the problem of
  biased datasets}. In \bibinfo{booktitle}{\emph{Proceedings of the 2019
  Conference of the North American Chapter of the Association for Computational
  Linguistics: Human Language Technologies, Volume 1 (Long and Short Papers)}}.
  \bibinfo{pages}{602--608}.
\newblock


\bibitem[\protect\citeauthoryear{Xu, Ba, Kiros, Cho, Courville, Salakhudinov,
  Zemel, and Bengio}{Xu et~al\mbox{.}}{2015}]%
        {xu2015show}
\bibfield{author}{\bibinfo{person}{Kelvin Xu}, \bibinfo{person}{Jimmy Ba},
  \bibinfo{person}{Ryan Kiros}, \bibinfo{person}{Kyunghyun Cho},
  \bibinfo{person}{Aaron Courville}, \bibinfo{person}{Ruslan Salakhudinov},
  \bibinfo{person}{Rich Zemel}, {and} \bibinfo{person}{Yoshua Bengio}.}
  \bibinfo{year}{2015}\natexlab{}.
\newblock \showarticletitle{Show, attend and tell: Neural image caption
  generation with visual attention}. In \bibinfo{booktitle}{\emph{International
  conference on machine learning}}. \bibinfo{pages}{2048--2057}.
\newblock


\bibitem[\protect\citeauthoryear{Zhang, Bargal, Lin, Brandt, Shen, and
  Sclaroff}{Zhang et~al\mbox{.}}{2018}]%
        {zhang2018top}
\bibfield{author}{\bibinfo{person}{Jianming Zhang}, \bibinfo{person}{Sarah~Adel
  Bargal}, \bibinfo{person}{Zhe Lin}, \bibinfo{person}{Jonathan Brandt},
  \bibinfo{person}{Xiaohui Shen}, {and} \bibinfo{person}{Stan Sclaroff}.}
  \bibinfo{year}{2018}\natexlab{}.
\newblock \showarticletitle{Top-down neural attention by excitation backprop}.
\newblock \bibinfo{journal}{\emph{International Journal of Computer Vision}}
  \bibinfo{volume}{126}, \bibinfo{number}{10} (\bibinfo{year}{2018}),
  \bibinfo{pages}{1084--1102}.
\newblock


\bibitem[\protect\citeauthoryear{Zhao, Wang, Yatskar, Cotterell, Ordonez, and
  Chang}{Zhao et~al\mbox{.}}{2019}]%
        {zhao2019gender}
\bibfield{author}{\bibinfo{person}{Jieyu Zhao}, \bibinfo{person}{Tianlu Wang},
  \bibinfo{person}{Mark Yatskar}, \bibinfo{person}{Ryan Cotterell},
  \bibinfo{person}{Vicente Ordonez}, {and} \bibinfo{person}{Kai-Wei Chang}.}
  \bibinfo{year}{2019}\natexlab{}.
\newblock \showarticletitle{Gender Bias in Contextualized Word Embeddings}. In
  \bibinfo{booktitle}{\emph{Proceedings of the 2019 Conference of the North
  American Chapter of the Association for Computational Linguistics: Human
  Language Technologies, Volume 1 (Long and Short Papers)}}.
  \bibinfo{pages}{629--634}.
\newblock


\bibitem[\protect\citeauthoryear{Zhao, Wang, Yatskar, Ordonez, and Chang}{Zhao
  et~al\mbox{.}}{2017}]%
        {zhao2017men}
\bibfield{author}{\bibinfo{person}{Jieyu Zhao}, \bibinfo{person}{Tianlu Wang},
  \bibinfo{person}{Mark Yatskar}, \bibinfo{person}{Vicente Ordonez}, {and}
  \bibinfo{person}{Kai-Wei Chang}.} \bibinfo{year}{2017}\natexlab{}.
\newblock \showarticletitle{Men Also Like Shopping: Reducing Gender Bias
  Amplification using Corpus-level Constraints}. In
  \bibinfo{booktitle}{\emph{Proceedings of the 2017 Conference on Empirical
  Methods in Natural Language Processing}}. \bibinfo{pages}{2979--2989}.
\newblock


\end{thebibliography}



\newpage
\appendix

\section{COCO-GB dataset}
\subsection{Gender Annotation} We show the gender word list in Tab.~\ref{tab:gender_words_list}. Gender words are selected based on the word frequency in COCO dataset. We delete the gender words that appear less than ten times in training data. Word "woman" and "man" are the most frequent gender-specific word and account for more than 60\% of the total gender-specific words.

\begin{table}[h]
    \centering
    \begin{tabular}{c|c}
    \toprule
        female word  ($g_w$) & \tabincell{c}{woman, women, girl, sister, \\ daughter, wife, girlfriend} 
        \\ \hline
        male word  ($g_m$) & \tabincell{c}{man, men, boy, brother, \\ son, husband, boyfriend} \\ \hline
        gender neutral word  ($g_n$) & people, person, human, baby\\
    \bottomrule
    \end{tabular}
    \caption{Gender words list. We label the caption based on the gender word appeared in the sentences.}
    \label{tab:gender_words_list}
\end{table}

\begin{figure*}
    \centering
    \includegraphics[width=1.0\linewidth]{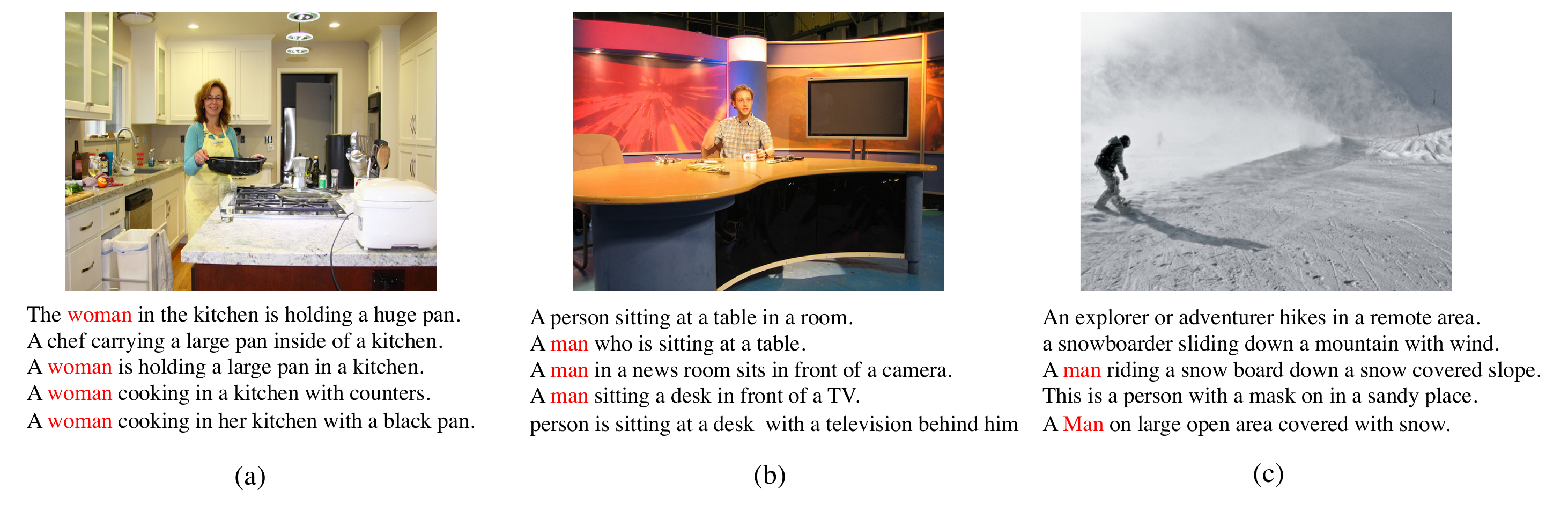}
    \vspace{-25pt}
    \caption{(a): An image is labeled as women. (b): An image is labeled as men. (c) An image is labeled as men. However, the gender evidence is actually occluded and our human evaluators label the image as "discard". }
    \label{fig:label_samples}
\end{figure*}

\noindent Some gender annotation examples are shown in Fig.~\ref{fig:label_samples}. We label an image as "women" when at least one sentence mentioned female words and label an image as "men" when at least one sentence mentioned male words. Images that both mention male words and female words are discarded.

We conduct human evaluations on our gender annotations with majority voting from 3 human evaluators. Each evaluator validates a total of 400 test samples, with 200 randomly sampled from each gender. Results show that our gender annotation achieves high precision, 92\% for women, and 95\% for men. Fig.~\ref{fig:label_samples} (c) shows an example of failure. The gender label based on captions is Men. However, evaluators label the image as "discard" since the gender is actually vague in the image. By analyzing the examples of failure, we observe that most conflicts occur when gender evidence is vague or occluded. In these cases, COCO annotators tend to provide gender identification based on context cues or social stereotypes.
\label{sec:label gender}

\subsection{Gender-Contex Joint Distribution}
\begin{figure}[h]
    \centering
    \includegraphics[width=1.0\linewidth]{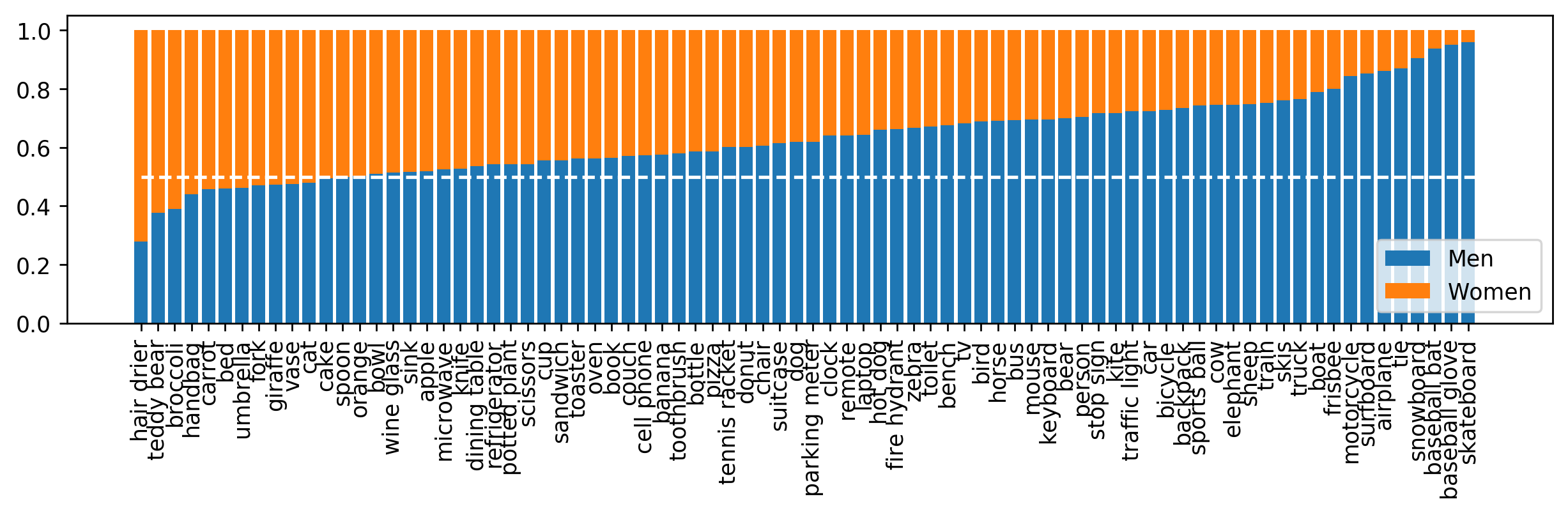}
    \caption{Gender-context distribution of COCO train}
    \label{fig:COCO training dataset}
\end{figure}

\begin{figure}[h]
    \centering
    \includegraphics[width=1.0\linewidth]{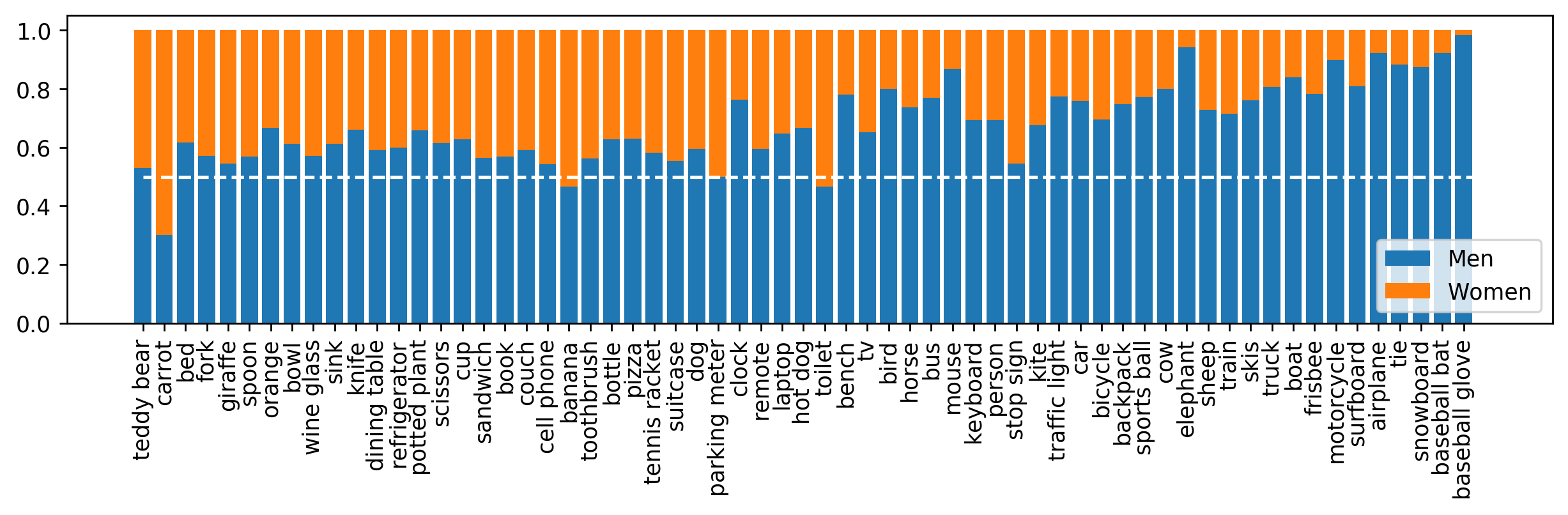}
    \caption{Gender-context distribution of original COCO test }
    \label{fig:COCO testing dataset}
\end{figure}

\begin{figure}[h]
    \centering
    \includegraphics[width=1.0\linewidth]{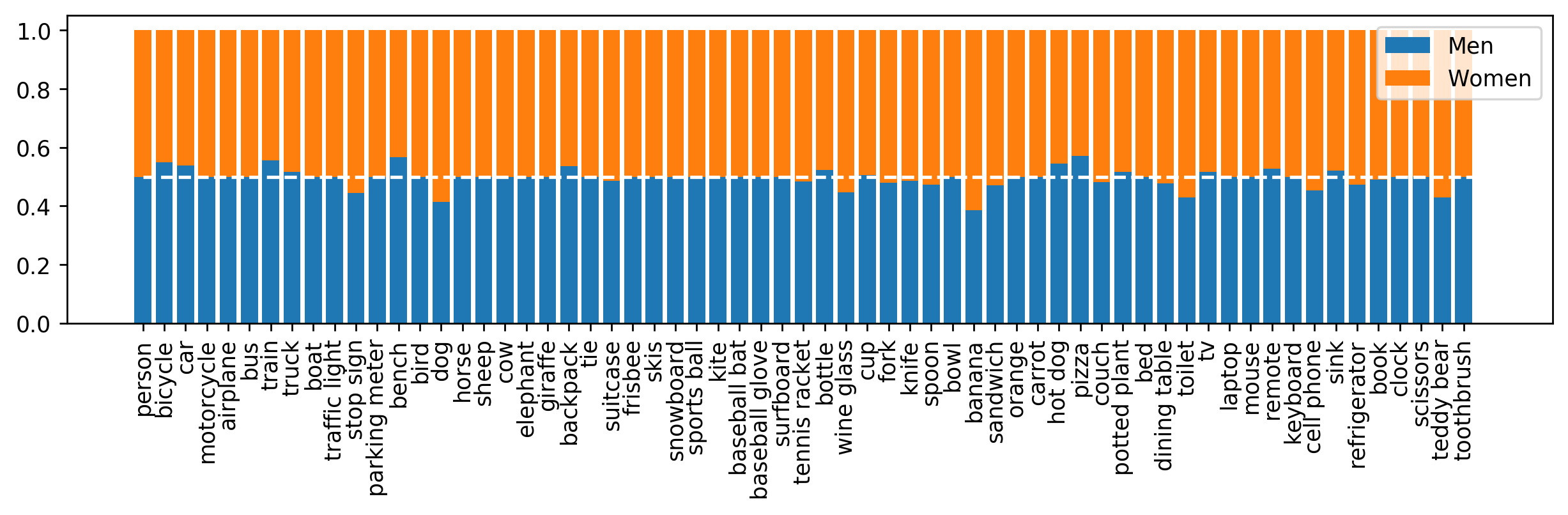}
    \caption{Gender-context distribution of COCO-GB V1 test }
    \label{fig:COCO V1 dataset}
\end{figure}

\begin{figure}[h]
    \centering
    \includegraphics[width=1.0\linewidth]{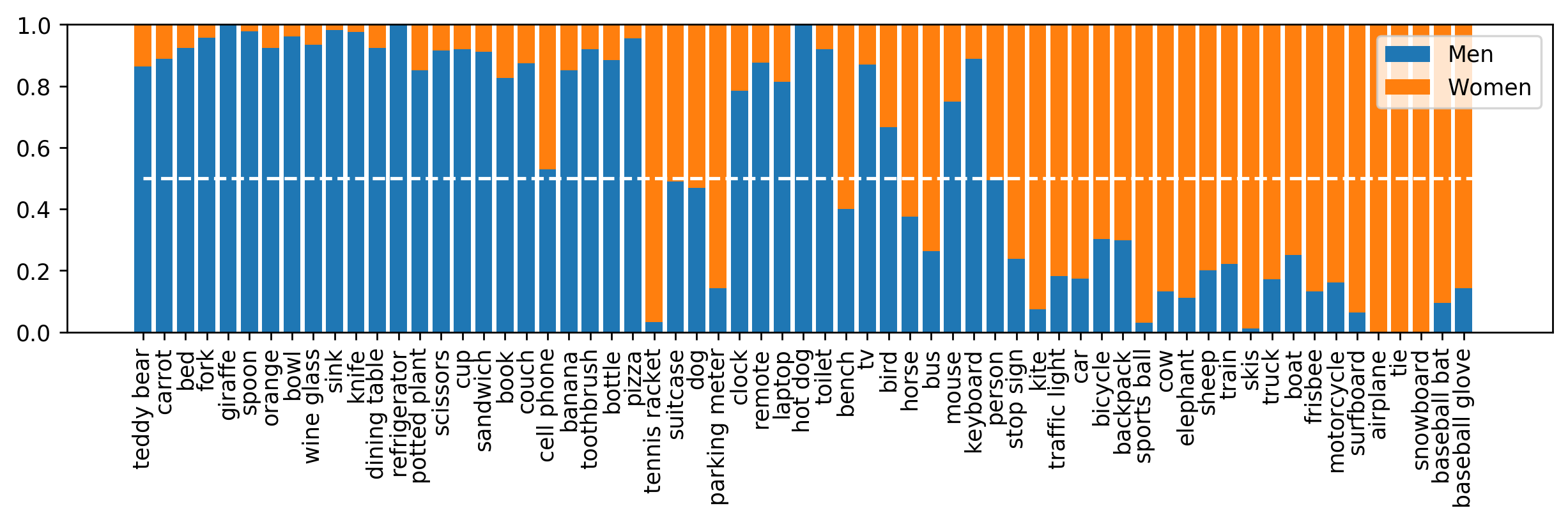}
    \caption{Gender-context distribution of COCO-GB V2 test }
     \label{fig:COCO V2 dataset}
\end{figure}

In Fig.~\ref{fig:COCO training dataset}-\ref{fig:COCO V2 dataset}, we show the gender-context joint distribution of COCO training dataset (Karpathy split), COCO testing dataset (Karpathy split), COCO-GV V1 secret testing dataset, COCO-GB V2 testing dataset. We list the objects in order of bias rate in COCO training dataset. For COCO-GB dataset, we choose 63 representative objects from a total of 80 objects and delete objects that do not occur in the corpus with sufficient frequency to be include. 

The Fig.~\ref{fig:COCO training dataset} shows that most objects in training data are more likely to co-occur with men. In Fig.~\ref{fig:COCO testing dataset}, we observe that a similar bias is also found in the test dataset. Hence directly evaluating the models on the biased test dataset might underestimate the gender bias learned by models. In comparison, we observe that COCO-GB V1 has a balanced gender-context joint distribution. Each object has an almost equal probability of occurring with men and women. COCO-GB V2 has a different distribution that gender-context joint distribution is opposite to the training set, which makes the gender prediction more difficult. Models relying on image context to provide gender identification will suffer from a huge gender prediction error on the anti-stereotypical test dataset.
\label{sec:coco_gender}

\begin{table*}
\begin{threeparttable}
\centering
\begin{tabular}{cccccccccc}
\toprule
\multicolumn{1}{c|}{\multirow{2}{*}{Model}} &  \multicolumn{1}{c|}{\multirow{2}{*}{C}} & \multicolumn{1}{c|}{\multirow{2}{*}{M}} & \multicolumn{3}{c|}{Woman} & \multicolumn{3}{c|}{Men} &
\multicolumn{1}{c}{\multirow{2}{*}{D}}\\
\cline{4-9} 
\multicolumn{1}{c|}{} & \multicolumn{1}{l|}{} & \multicolumn{1}{l|}{} & \multicolumn{1}{l|}{correct} & \multicolumn{1}{l|}{wrong} & \multicolumn{1}{l|}{neutral} & \multicolumn{1}{l|}{correct} & \multicolumn{1}{l|}{wrong} & 
\multicolumn{1}{l|}{neutral} \\ \hline
Baseline & 98.2 & 27.2 & 51.6 & 28.3 & 20.1 & 77.9 & \textbf{4.9} & 17.1 & 0.094  \\
Balanced & 97.5 & 27.3 & 57.9 & 25.5 & 26.6 & 71.1 & 11.5 & 17.4  & 0.034 \\
UpWeight-10 & 95.8 & 26.9 & \textbf{72.2} & 26.1 & 1.7 & \textbf{86.1} & 11.7 & 2.1 & 0.023  \\
PixelSup & 96.8 & 27.1 & 54.2 & 25.1 & 20.5 & 76.4 & 6.2 & 17.2  & 0.062 \\
\hline
GAIC & 97.8 & 26.9 & 67.1 & 18.0 & 14.9 & 68.9 & 10.7 & 20.3 & 0.008 \\
GAIC$_{es}$ & 98.1 & 27.0 & 69.1 & \textbf{15.2} & 15.7 & 71.4 & 8.1 & 20.5  & \textbf{0.007} \\
\bottomrule
\end{tabular}
\caption{Gender bias analysis on COCO-GB V2 split.}
\end{threeparttable}
\label{tab:COCO-GB V2}
 \vspace{-20pt}
\end{table*}

\section{More on Implementation Details}
\textbf{Benchmarking Baselines:} All the baselines mentioned in Sec.~\ref{sec:benchmarking} use visual features extracted from the fourth layer of ResNet-101. All baselines except for NBT, TopDown, Att, and AdaptAtt are implemented in a same open-source framework \footnote{https://github.com/ruotianluo/self-critical.pytorch}.  We directly use the caption results from a web source \footnote{https://github.com/LisaAnne/Hallucination} to measure gender prediction performance. For NBT, TopDown, Att, and AdaptAtt models, we implement models based on the paper and make sure that the model's caption score is close to the results reported in work. For all models, we evaluate caption quality by the official COCO evaluation tool \footnote{https://github.com/tylin/coco-caption}.

\vspace{2pt}
\noindent\textbf{Debiasing Baselines:} 
Here we discuss the implementation details of baselines mentioned in Sec.~\ref{sec:baseline of debiasing}. For baseline Balanced, we randomly select a subset from the original training set which contains 4,000 images for each gender. Balanced model is obtained by fine-tuning the Att model on this selected dataset with 5 epochs. For UpWeight model, we first train the Att model with normal loss. Then we upweight the loss value of gender words and continue to train the model for another 1 epoch. For model PixelSup, we first train the Att model with normal loss. Then we utilize Eq.~\ref{eq:mask} to further fine-tune the attention learning process on the 10\% extra supervision data with one more epoch.

\vspace{2pt}
\noindent\textbf{GAIC Model:} 
  For GAIC model, we adopt the two-streams pipeline to train the model. In the experiments, we find $\mu=0.1$ performs best, large $\mu$ value will influence caption qualiy, small $\mu$ value cannot efficiently mitigate gender bias. For GAIC$_{es}$ model, we fine-tune the dataset with extra human instance segmentation annotations, and set $\mu=0.1$ and $\eta=0.05$. Like $\mu$ value, we empirically find that large  $\eta$ value will significantly impact caption quality. Hence we choose a relatively small value $\eta=0.05$ to train GAIC$_{es}$ model.

\section{Experimental Results on COCO-GB V2}
Compared to COCO-GB V1, all baselines in COCO-GB V2 obtain a higher error rate, especially for women (average increase of 3.25\%). GAIC model improves the gender prediction accuracy of woman from 51.6\% to 67.1\% and reduces the error rate of women from 28.3\% to 18.0\%. Although the UpWeight-10 model obtains the highest accuracy of both women and men, it causes an unacceptably high error rate for two genders. There is no substantial difference between the Balanced model and baseline model, a similar trend has been found in COCO-GB V1. Compared to GAIC, GAIC$_{es}$ obtains consistently better performance. For fairness evaluation, we compare different model’s gender divergence. GAIC and GAIC$_{es}$ obtain the lowest divergence, which indicates that models treat each gender in a fair manner. For caption quality, GAIC and GAIC$_{es}$ only cause a minor performance drop compared to the baseline model, which proves the robustness of the self-exploration training strategy.
\label{sec:COCO-GB V2 result}

\section{More Qualitative Results}
We show more visualizing attention results in Fig.~\ref{sec:More Qualitative Results}. We observe that the baseline Att model tends to utilize context features to predict gender and thus makes incorrect gender prediction, e.g., predicting a woman as a man based on a tie. In comparison, GAIC and GAIC$_{es}$ learn to use the features on the regions of the described person for gender prediction. The result also shows that when gender evidence is vague, GAIC and GAIC$_{es}$ model tend to use neutral gender words, such as "person" to describe the person.

\section{More on Ethics concern}
In the COCO dataset, people in the images are not asked to identify their biological gender. Thus we emphasize that we are not classifying biological sex or gender identity, but rather outward gender appearance. In this work, we follow the settings used in previous work to define three gender categories: male, female, and gender-neutral (when the image is blurry or the gender in the image is ambiguous) based on visual appearance. Gender labels are determined based on the annotators' descriptions.
\begin{figure*}[h]
    \centering
    \includegraphics[width=0.9\linewidth]{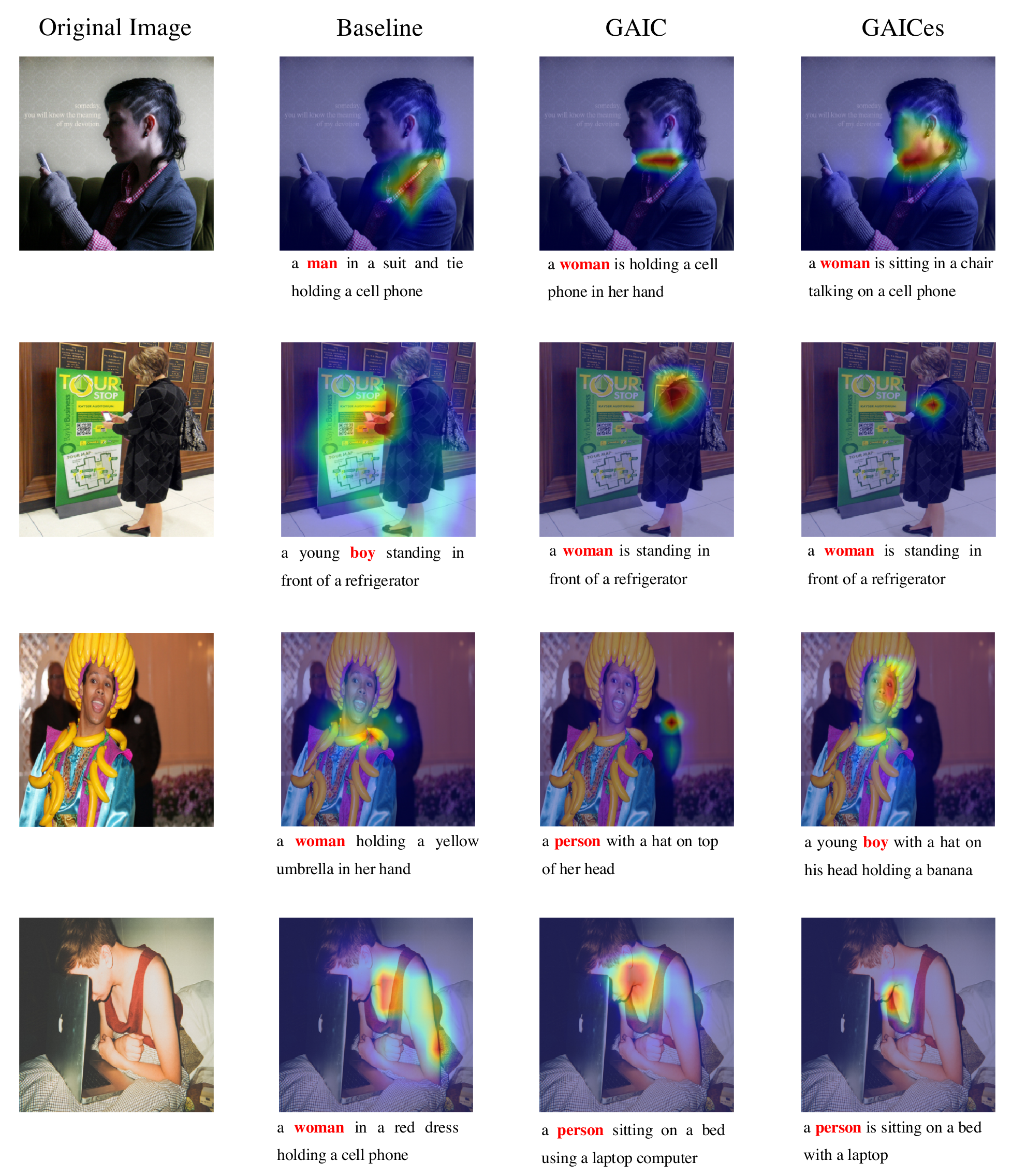}
\caption{Qualitative comparison of baselines and our proposed model. At the top, we show success cases that our proposed modes correctly predict the gender and utilize proper visual evidence. The bottom case shows that when gender evidence is vague, our model tends to use neutral gender words, such as "person" to describe the gender of the person.}
\end{figure*}
\label{sec:More Qualitative Results}
\end{document}